\documentclass[sigplan,screen,nonacm]{acmart}
\AtBeginDocument{%
  }

\usepackage{algorithm}
\usepackage[noend]{algpseudocode}
\usepackage{xcolor}
\usepackage{tikz}
\usepackage{amsmath}
\usepackage{graphicx}
\usepackage{booktabs}
\usepackage{hyperref}
\usepackage{enumitem}
\usepackage{xspace}
\usepackage{soul}
\usepackage{verbatim}
\usepackage{multirow}
\usepackage{pifont}
\usepackage{makecell}
\usepackage{listings}

\newcommand{\stitle}[1]{\noindent \textbf{{#1}}}

\newcommand{\lhy}[1]{{#1}}

\newcommand{\pivotrl}{{\textsf{PivotRL}}\xspace}

\newcommand{\timing}[2]{$#1\,{\scriptstyle \pm #2}$}
\definecolor{commentblue}{RGB}{0,0,139}

\newcommand{\myvspace}[1]{\vspace{#1}}

\newcommand\footnotewithoutmarker[1]{%
  \begingroup
  \renewcommand\thefootnote{}\footnote{#1}%
  \addtocounter{footnote}{-1}%
  \endgroup
}

\begin{document}

\title{Efficient Resource-Shiftable RL Post-training with Generative Rewards}

\pagestyle{plain}


\author{Yanfeng Zhao$^{*\S}$, Haoyang Li$^{*\dag}$, Sheng Lin$^{\dag}$, Yuming Zhou$^{\dag}$, Jiawei Jiang$^{\ddag}$, Bin Cui$^{\dag}$, Fangcheng Fu$^{\S}$\\
$^{\S}$Shanghai Jiao Tong University, $^{\dag}$Peking University, $^{\ddag}$Central China Normal University\\
$^{*}$Equal Contribution}


\begin{abstract}
\lhy{Reinforcement learning (RL) post-training with generative reward models (GenRM) has become a central recipe for improving large language models (LLMs), and it is also increasingly resource-intensive: policy rollout, GenRM-based reward evaluation, and policy training all compete for GPU resources. However, existing colocated and disaggregated systems rely on static resource allocation for these roles. Such allocation cannot adapt as each role's workload changes during RL and therefore leaves a role's resources underutilized.}

\lhy{We present \pivotrl, an RL post-training system that dynamically shifts resources among different roles according to their runtime demands. \pivotrl introduces a resource-shiftable runtime built on a Global Parameter Storage, together with a unified mechanism that efficiently shifts role-specific context. A shift-aware scheduler enables sequence migration upon a shift, while a planner considers the aggregate throughput benefit and the shift overhead to determine when and how to shift. We evaluate \pivotrl on a 128-GPU cluster using four model groups spanning dense and mixture-of-experts models. \pivotrl enables resource shifting within seconds and improves end-to-end throughput by 1.81--2.05$\times$ over colocated deployment and 1.50--1.95$\times$ over disaggregated deployment, while preserving RL convergence.}
\footnotewithoutmarker{Contacts: Yanfeng Zhao (\mbox{victor\_yfzhao@sjtu.edu.cn}), Haoyang Li (\mbox{lihaoyang@stu.pku.edu.cn}), and Fangcheng Fu (\mbox{ccchengff@sjtu.edu.cn}).}
\end{abstract}

\maketitle

\section{Introduction}
\label{sec:intro}

\begin{figure}[t]
\centering
\includegraphics[width=\columnwidth]{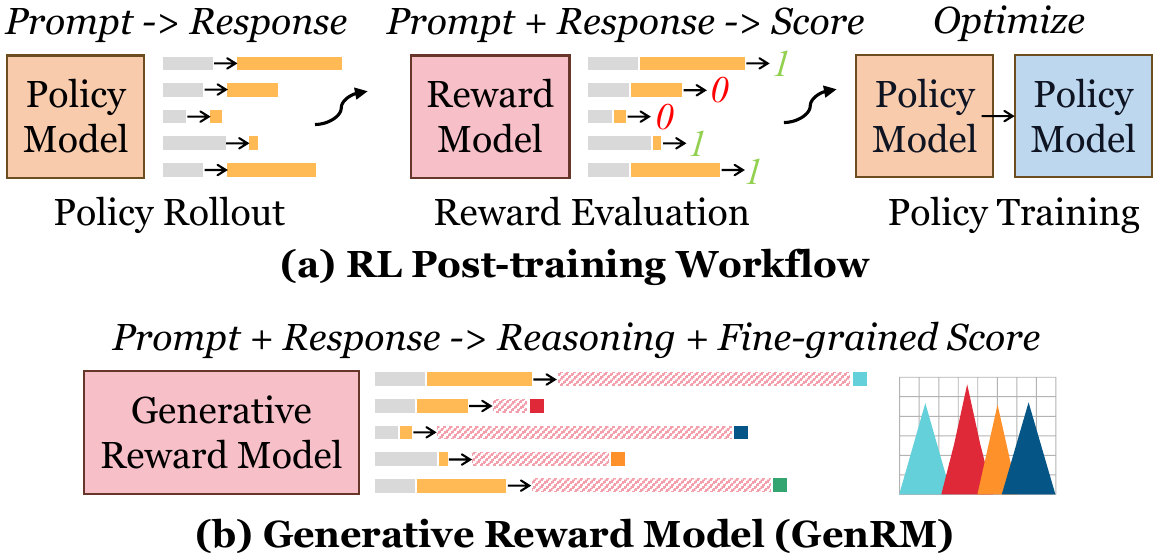}
\myvspace{-21pt}
\caption{\small{\lhy{(a) Three stages of a typical RL post-training step. (b) Generative reward models (GenRMs) provide more fine-grained feedback by reasoning over the responses of the policy model.}}}
\label{fig:workflow}
\myvspace{-10pt}
\end{figure}

\begin{figure*}[t]
\centering
\includegraphics[width=\linewidth]{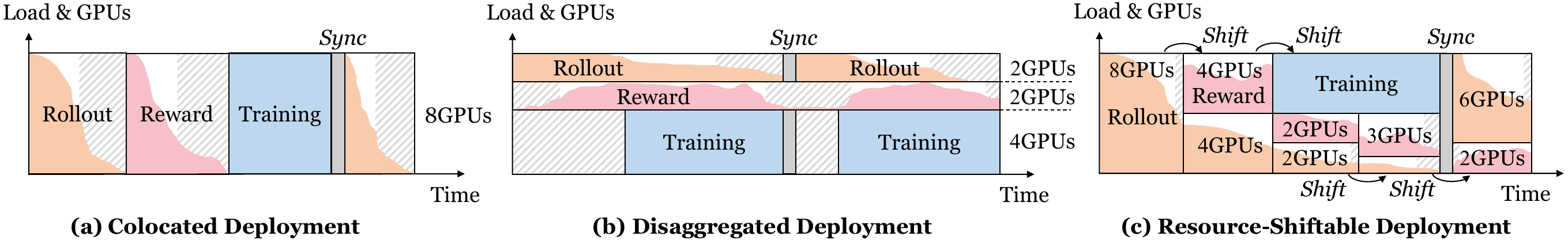}
\myvspace{-22pt}
\caption{\small{\lhy{Comparison of deployment paradigms. Both (a) colocated and (b) disaggregated deployments statically allocate resources, while (c) resource-shiftable deployment dynamically reallocates GPUs to match workload demand. Different colors indicate instantaneous stage load, and shaded regions indicate throughput headroom, where workload is substantially below the processing capacity of allocated GPUs.}}}
\label{fig:compare}
\myvspace{-10pt}
\end{figure*}

\lhy{Reinforcement learning (RL) has become an essential step in post-training large language models (LLMs), driving recent advances in their reasoning capabilities~\cite{scale_rl_1, scale_rl_2, rl_reasoning_3}. As shown in Figure~\ref{fig:workflow}(a), a typical RL step consists of three stages: the policy model (i.e., the LLM) generates responses to a batch of prompts (policy rollout), the responses are scored to produce reward signals (reward evaluation), and the signals are used to update the policy (policy training).}

\lhy{An increasingly prominent trend in this pipeline is the shift in how rewards are produced. Early RL recipes rely on rule-based verifiable rewards, which compare a response against a ground-truth answer and emit a binary correctness signal~\cite{rlvr_1, rlvr_2}. While effective for tasks with well-defined answers, such rewards are unavailable for open-ended tasks such as long-form question answering, creative writing, and mathematical reasoning without golden answers. Recent work therefore complements verifiable rewards with \emph{generative rewards}~\cite{generative_verifier, think_rm, gen_rm, rm_r1, r_align, RM-NLHF}. As shown in Figure~\ref{fig:workflow}(b), a generative reward model (GenRM), itself an LLM, reasons over the response before producing a reward. Compared with $0/1$ correctness signals or other outcome-based rewards, GenRMs can provide more fine-grained feedback through soft rewards, such as probabilistic or continuous scores that capture the degree of correctness or preference, thereby providing richer training signals~\cite{verify, rlaif}. Consequently, GenRMs are increasingly adopted in the post-training of frontier LLMs, including DeepSeek-R1~\cite{deepseek_r1}, GLM-5~\cite{glm5}, and Kimi-K3~\cite{kimi-k3}.}

\lhy{This trend makes RL post-training substantially more resource-intensive. With GenRMs, reward evaluation is no longer a lightweight scalar computation, but a full LLM inference workload. As a result, all three RL stages now require substantial GPU resources. 
How to deploy these stages on a given cluster therefore becomes a critical systems question.}

\lhy{Existing RL systems answer this question with one of two deployment paradigms: (1) \emph{Colocated deployment}~\cite{verl, openrlhf, trl, slime} (Figure~\ref{fig:compare}(a)), in which the three stages share the whole cluster and run sequentially, swapping models in and out of GPU memory at stage boundaries. (2) \emph{Disaggregated deployment}~\cite{stream_rl, llama_rl, areal, roll_flash, rhy_rl, laminar, async_rlhf, asyncflow,staleflow} (Figure~\ref{fig:compare}(b)), in which the cluster is partitioned into disjoint GPU groups, one per stage. These stages then execute asynchronously.}

\lhy{However, as Figure~\ref{fig:compare} illustrates, the workload of each stage varies continuously within a step. For rollout and reward evaluation, the number of active sequences\footnote{A sequence refers to a generation of the policy model or the GenRM.} decreases as the step progresses, while training cannot proceed until a full batch of rewards has accumulated. Under colocation, rollout and reward evaluation therefore have substantial throughput headroom in the tail of every batch. Under disaggregation, rollout suffers from the same tail effect, reward evaluation fluctuates with rollout's instantaneous production rate, and training GPUs remain idle while waiting for the next batch. As a result, both colocated and disaggregated deployments yield suboptimal end-to-end throughput.}

\lhy{The root cause is static resource allocation: regardless of whether colocation or disaggregation is used, each stage remains bound to a fixed set of GPUs throughout RL post-training. This static allocation cannot adapt to the dynamic workload demands across stages. When one stage has substantial throughput headroom, its spare resources cannot be used by another stage experiencing higher demand.}

\lhy{We therefore argue, as illustrated in Figure~\ref{fig:compare}(c), that resolving this mismatch calls for a more flexible deployment paradigm in which resources can shift across stages at runtime, a paradigm we refer to as \textit{resource-shiftable deployment}. Under this paradigm, GPUs are dynamically reallocated according to instantaneous workload. For example, as rollout's load decreases, its spare GPUs can shift to reward, and when rollout becomes the bottleneck, GPUs can shift back. Likewise, once training completes, GPUs that would otherwise remain idle can be immediately reallocated to rollout or reward. By dynamically matching resource capacity with workload demand, resource-shiftable deployment can better utilize the resources and improve end-to-end throughput.
}

\lhy{To realize this paradigm, we introduce \pivotrl, a system that accelerates LLM RL with generative rewards through resource shifting. It entails three key innovations. (1) \textit{Supporting resource shifting.} Each resource (i.e., a GPU) holds a set of model parameters (policy model or GenRM) and is bound to a role (i.e., an RL stage), so shifting a resource means shifting both the parameters and the execution context of the role. \pivotrl addresses this with a Global Parameter Storage that lets any GPU fetch the parameters of any model on demand, together with a unified role-shifting mechanism, enabling smooth resource shifting between any models and any roles. (2) \textit{Rescheduling workload after a shift.} A shift changes the resources a role owns, so the role's workload must be rescheduled under its new resource allocation. \pivotrl designs a shift-aware scheduler that supports shift-time sequence migration across replicas, maximizing the role's throughput under the new allocation. (3) \textit{Deciding when and how to shift.} \pivotrl introduces a planner to determine when to trigger a shift, which role should donate resources to another, and exactly which resources to shift. It evaluates candidate shift plans based on their expected throughput benefit and shifting overhead, and executes the most profitable plan.}

\lhy{In summary, this paper makes the following contributions:}                                               
\begin{itemize}[noitemsep, topsep=0pt, parsep=0pt, partopsep=0pt, leftmargin=*]
\item \lhy{To the best of our knowledge, we are the first to systematically study LLM RL with generative rewards and identify static resource allocation as a key system bottleneck.}                                                
\item \lhy{We propose resource-shiftable deployment and realize it in \pivotrl through three innovations: a resource-shiftable runtime built on a Global Parameter Storage with a unified role-shifting mechanism, a shift-aware scheduler that reschedules workload across a role's replicas, and a planner that decides when and how to shift.}                                           
\item \lhy{We evaluate \pivotrl on a 128-GPU cluster across four model groups spanning dense and mixture-of-experts models. \pivotrl completes a resource shift within seconds and improves end-to-end throughput by $1.81$--$2.05\times$ over colocated deployment and $1.50$--$1.95\times$ over disaggregated deployment, while preserving RL convergence.}
\end{itemize}

\section{\lhy{Preliminaries}}
\label{sec:background}

\subsection{\lhy{RL Post-training with Generative Reward Models}}

\lhy{As shown in Figure~\ref{fig:workflow}(a), an RL post-training pipeline~\cite{ppo, grpo, dapo} consists of three major stages: policy rollout, reward evaluation, and policy training. During rollout, the policy model (i.e., the model being post-trained) generates responses to a set of prompts. The generated responses are then evaluated by a reward model (RM) to produce reward signals, which are subsequently used to update the policy model. The updated model parameters are then synchronized back to the rollout stage, completing one step of the RL loop.}  

\lhy{Recently, generative reward models (GenRMs) have been increasingly used to provide richer reward signals~\cite{generative_verifier, think_rm, gen_rm, rm_r1, r_align, RM-NLHF}. As shown in Figure~\ref{fig:workflow}(b), a GenRM itself is an LLM that reasons over the rollout response before producing its reward. This makes RL increasingly resource-intensive, as all three RL stages require substantial GPU resources. A fundamental system's question is therefore how to efficiently deploy these stages on a given GPU cluster.}

\subsection{\lhy{Different Deployment Paradigms}}

\lhy{Existing systems mainly adopt colocated~\cite{verl, openrlhf, trl, slime} or disaggregated~\cite{stream_rl, llama_rl, areal, roll_flash, rhy_rl, laminar, async_rlhf, asyncflow,staleflow} deployment.}

\stitle{\lhy{Colocated deployment.}}
\lhy{As illustrated in Figure~\ref{fig:compare}(a), colocated deployment shares the entire GPU cluster across all three stages, which execute sequentially. 
When transitioning between stages, the current model is offloaded to CPU memory and the next model is loaded onto the GPUs.}

\stitle{\lhy{Disaggregated deployment.}}
\lhy{As illustrated in Figure~\ref{fig:compare}(b), disaggregated deployment statically partitions the GPU cluster among the three stages, which execute concurrently and asynchronously. 
Once a rollout sequence finishes, the reward stage consumes it immediately to produce a reward, while the training stage performs a policy update only after sufficient rewards accumulate to form a training batch.}

\begin{figure}
    \centering
    \includegraphics[width=\linewidth]{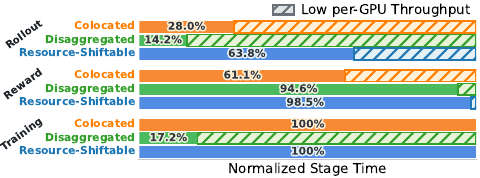}
    \myvspace{-20pt}
    \caption{\small{
    \lhy{The per-GPU throughput under different deployment paradigms on 32 GPUs, using Qwen2.5-7B as the policy model and Qwen3-30B-A3B as the GenRM. Shaded regions indicate the fraction of execution time spent at low per-GPU throughput.}
    }}
    \label{fig:low_util}
    \myvspace{-10pt}
\end{figure}

\lhy{However, both paradigms statically allocate GPUs across stages, causing resource allocation to mismatch dynamic workload demands and leaving substantial throughput headroom. We therefore introduce resource-shiftable deployment, as illustrated in Figure~\ref{fig:compare}(c), which dynamically shifts resources across stages to match changing workload demands.} 

\lhy{In Figure~\ref{fig:low_util}, we quantify the throughput headroom of different deployment paradigms using real RL workloads. Specifically, we measure the fraction of time each stage spends at a low per-GPU throughput, defined as throughput below 20\% of its peak. We find that both colocated and disaggregated deployments leave substantial throughput headroom in the rollout stage. While disaggregated deployment can significantly reduce the throughput headroom of the reward stage, it comes at the cost of training idleness. In contrast, resource-shiftable deployment (implemented in \pivotrl) reduces throughput headroom across all three stages, resulting in higher end-to-end performance.}

\section{\lhy{Challenges}}
\label{sec:challenge}

\lhy{Realizing resource-shiftable deployment, and turning it into actual end-to-end throughput gains, however, is far from straightforward. It involves three core challenges.} 

\stitle{\lhy{C1: How to support resource-shiftable deployment.}}
\lhy{Consider a single resource (i.e., a GPU). At any moment, it holds a set of model parameters and carries a \emph{role}, defined as the RL stage to which it is currently assigned. Beyond the parameters, a role also requires its own execution context on the device, such as the CUDA Graphs captured for rollout or the optimizer states and NCCL communication groups established for training. Shifting a resource from one RL stage to another therefore includes shifting both its parameters and the additional context tied to its role. This is particularly challenging in RL post-training, where a single run spans two models (the policy model and the GenRM) and three roles (rollout, reward, and training), yielding a wide variety of shift cases that are impractical to handle in an ad hoc manner. Supporting flexible resource shifting therefore calls for a unified system infrastructure and shifting mechanism.}

\stitle{\lhy{C2: How to redesign the scheduler under resource shifting.}}
\lhy{Every shift carries an immediate side effect: the affected roles gain or lose resources. Newly acquired resources are useful only if part of the workload is rescheduled onto them, while removed resources require rescheduling the affected workload elsewhere. This rescheduling directly determines the role's attainable throughput under the new allocation. Existing schedulers in RL systems~\cite{verl, areal, slime, roll}, however, typically support only dispatch (i.e., assigning sequences to replicas), not rescheduling. The scheduler must therefore be redesigned under resource shifting.}

\stitle{\lhy{C3: How to make shift decisions.}}
\lhy{While rescheduling is a consequence of resource shifting, deciding how to shift resources is a first-class problem. A single shift decision must determine when to trigger the shift, which role donates resources, which role receives them, and exactly which resources to move. These choices are tightly coupled, and shifting is not free: parameter movement and role context reconstruction both incur overhead. The benefit of a shift therefore depends on its timing, its plan, and its associated overhead, making it a complex joint optimization problem. Formulating this optimization and solving it with an efficient algorithm is the final challenge we must address.}

\section{The Designs and Overview of \pivotrl}
\label{sec:overview}

\begin{figure}
    \centering
    \includegraphics[width=\linewidth]{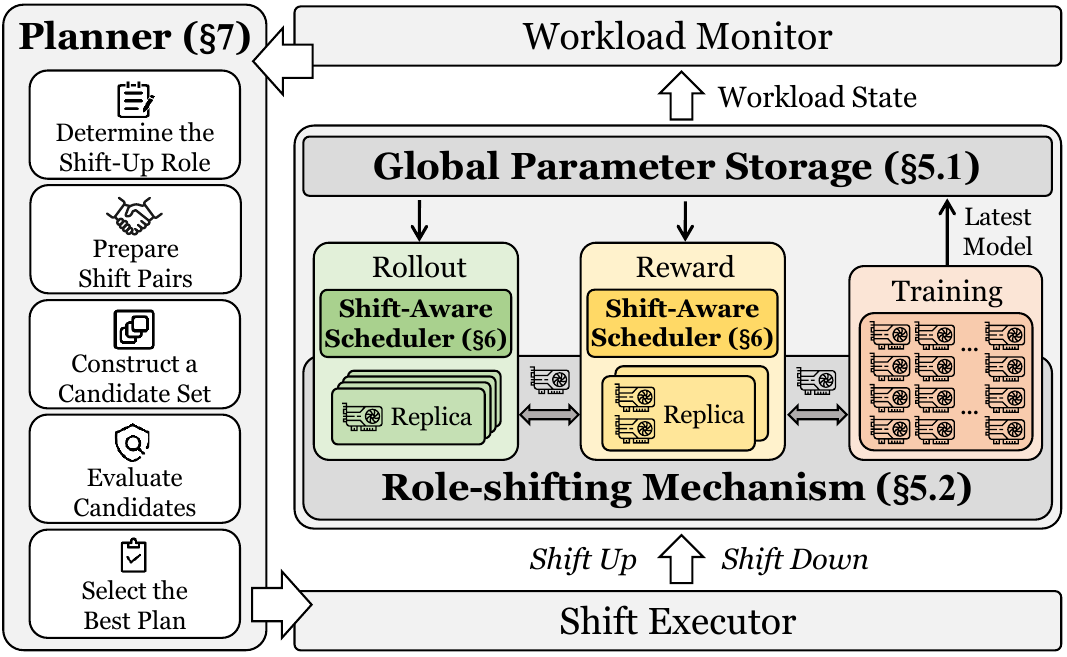}
    \myvspace{-20pt}
    \caption{\small{\lhy{Overview of \pivotrl, which dynamically shifts resources across RL stages via a resource-shiftable runtime, shift-aware schedulers, and a planner with monitor and executor.}}}
    \label{fig:overview}
    \myvspace{-10pt}
\end{figure}

\lhy{We introduce \pivotrl, an LLM RL system that dynamically shifts resources across RL stages so that resource capacity tracks workload demand, thereby improving throughput. As shown in Figure~\ref{fig:overview}, beyond the resources (i.e., GPUs) serving the three RL stages, \pivotrl introduces a set of additional components, each addressing one of the challenges identified in \S\ref{sec:challenge}. \ding{182} A system runtime makes resource-shiftable deployment possible (\S\ref{sec:elasticity}), which consists of a Global Parameter Storage (GPS) infrastructure (\S\ref{subsec:ps}) and a role-shifting mechanism (\S\ref{subsec:shift}). \ding{183} A shift-aware scheduler migrates workload after a resource shift (\S\ref{sec:rebalance}), which is adopted for both the rollout and reward evaluation stages. \ding{184} A planner produces the best shift plan (\S\ref{sec:planner}), paired with a monitor that tracks workload state and a shift executor that carries the plan out. We now walk through the core design of each component.}

\stitle{\lhy{Resource-Shiftable runtime (\S\ref{sec:elasticity}).}}
\lhy{To enable resource-shiftable deployment (\textbf{C1}), the system runtime needs to address two fundamental requirements: shifting model parameters and shifting the additional role context. For the former, \pivotrl adopts a Global Parameter Storage (GPS), which pools the CPU memory of the entire GPU cluster into a distributed store holding the parameters of all models (i.e., the policy model and the GenRM). Any GPU can then fetch whatever parameters it requires, making a shift possible in the first place (\S\ref{subsec:ps}). For the latter, \pivotrl employs a unified role-shifting mechanism that distinguishes stateless from stateful data and decomposes a shift into two steps: a \textit{shift down} of the old role followed by a \textit{shift up} of the new one, while allowing each role to shift at its own granularity (\S\ref{subsec:shift}).}

\stitle{\lhy{Shift-Aware scheduler (\S\ref{sec:rebalance}).}}
\lhy{
To redesign the scheduler under resource shifting (\textbf{C2}), we build on the existing abilities of the rollout and reward schedulers to continuously dispatch sequences, and augment it with shift-time handling so that rescheduling is naturally absorbed by the scheduler itself.\footnote{Training has no scheduler: it operates on the entire workload and requires no sequence-level scheduling.}
The scheduler maintains a shared waiting queue and models per-replica throughput to continuously dispatch sequences using a throughput-maximizing mechanism (\S\ref{subsec:queue}). 
A \textit{shift down} or \textit{shift up} is then straightforward: the scheduler aborts the sequences needing migration and returns them to the shared queue, which re-dispatches them through the same throughput-maximizing mechanism (\S\ref{subsec:migration}).
}

\stitle{\lhy{Planner (\S\ref{sec:planner}).}}
\lhy{To determine when and how to shift resources (\textbf{C3}), \pivotrl employs a planner with a five-step procedure that constructs a shift plan based on the current workload state (\S\ref{subsec:planner overview}). The procedure first determines the \textit{shift-up} role and prepares shift pairs. It then constructs candidate shift plans from these pairs using a dynamic programming algorithm (\S\ref{subsec:candidate condtruct}), and selects the most beneficial candidate by jointly considering its throughput gain and penalty (\S\ref{subsec:benefit evaluation}).}

\stitle{\lhy{Workflow.}}
\lhy{Putting these components together, \pivotrl operates as follows. The monitor continuously collects the workload state of different roles and reports it to the planner. The planner makes a shift plan, and the shift executor realizes it by invoking \textit{shift down} and \textit{shift up}, in cooperation with GPS. Once a shift completes, the shift-aware scheduler of the affected role reschedules workload across its replicas.}

\section{\lhy{Resource-Shiftable Runtime}}
\label{sec:elasticity}

\lhy{In this section, we present our system runtime that supports resource-shiftable deployment. We begin by introducing the Global Parameter Storage (GPS), which offers an interface for accessing model parameters (\S\ref{subsec:ps}). Leveraging this foundation, we propose a unified role-shifting mechanism (\S\ref{subsec:shift}).}

\subsection{\lhy{Global Parameter Storage}}
\label{subsec:ps}

\begin{figure*}
    \centering
    \includegraphics[width=\linewidth]{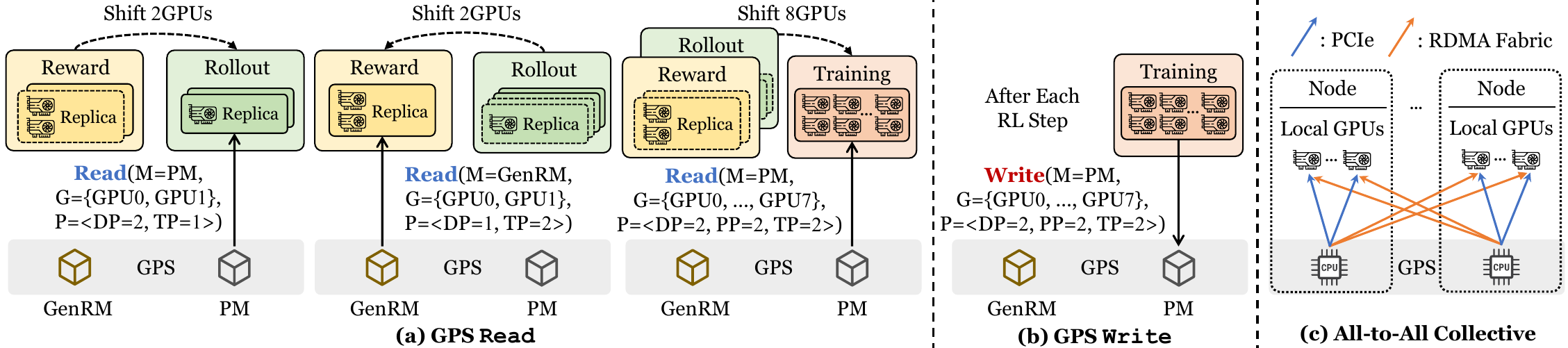}
    \myvspace{-23pt}
    \caption{\small{\lhy{The GPS interface. PM: policy model. DP/PP/TP: data/pipeline/tensor parallelism. (a) \texttt{Read}: after a resource shift, newly added resources fetch the required model parameters from GPS and partition them according to the parallelism strategy. (b) \texttt{Write}: after each RL step, training GPUs update GPS with the latest policy model parameters. (c) Both \texttt{Read} and \texttt{Write} leverage all-to-all collectives.
    }}}
    \label{fig:gps}
    \myvspace{-5pt}
\end{figure*}

\stitle{\lhy{Design rationale.}}
\lhy{To realize unified parameter shifting, the most straightforward approach is to transfer the required parameters directly from peer GPUs. However, this presupposes that the target parameters already reside somewhere in the cluster, an assumption that breaks once a role is shifted down entirely, leaving no replica on any GPU. An alternative is to reload the parameters from disk, which guarantees availability but incurs prohibitive I/O overhead.}

\lhy{We therefore mediate parameter transfer through CPU memory. Specifically, we maintain additional copies of the parameters for all models (e.g., the policy model and GenRM) in CPU memory across the cluster, forming the Global Parameter Storage (GPS). GPS leverages idle CPU memory already available in the cluster, without introducing external storage. It attains both availability and efficiency: parameters remain accessible regardless of the current GPU allocation, while the high I/O cost of a disk-based approach is avoided.}

\stitle{\lhy{Unified interface.}}
\lhy{GPS exposes exactly two parameter operations, \texttt{Read} and \texttt{Write}:}
\begin{lstlisting}[
basicstyle=\ttfamily\small,
keywordstyle=\bfseries,
commentstyle=\itshape,
xleftmargin=0em,
aboveskip=0.5em,
belowskip=0.5em
]
Read(M, G, P) # Model, GPUs, Parallelism
Write(M, G, P) # Model, GPUs, Parallelism
\end{lstlisting}
\lhy{$M$ denotes the target model, $G$ the set of GPUs that receives or sources its parameters, and $P$ the parallelism strategy under which $M$ is laid out on $G$.}

\lhy{These two operations suffice for all parameter movements induced by resource shifts in RL. As shown in Figure~\ref{fig:gps}(a), whenever resources are shifted, the newly assigned GPUs simply \texttt{Read} the model they are about to serve: rollout resources \texttt{Read} the policy model, reward resources \texttt{Read} GenRM, and training resources \texttt{Read} the policy model. All three cases invoke the same \texttt{Read} operation, independent of the role (i.e., RL stage) and of how resources are shifted, rendering GPS fully general to arbitrary resource shifts.} 

\lhy{Meanwhile, \texttt{Write} is invoked only when the contents of GPS need to be updated. During RL post-training, GenRM remains frozen, so \texttt{Write}$(M{=}\text{GenRM}, \ast)$ is never invoked. The policy model, however, is updated after each RL training stage. As illustrated in Figure~\ref{fig:gps}(b), the training stage commits the updated weights to GPS with a single \texttt{Write}, so all subsequent \texttt{Read}s can observe the latest policy.}

\stitle{Implementation.}
\lhy{Because different choices of $P$ induce different parameter partitionings across GPUs, while GPS stores parameters using its own fixed partitioning across CPUs, \texttt{Read} and \texttt{Write} are fundamentally a re-partitioning between CPUs and a set of GPUs. As shown in Figure~\ref{fig:gps}(c), following prior work~\cite{hotspa, hydraulis, hetu}, we realize this re-partitioning with an all-to-all collective. For engineering details of the parameter partitioning in GPS and the all-to-all implementation, we refer interested readers to Appendix~\ref{app:gps} for details.}

\stitle{\lhy{Overhead and scalability.}}
\lhy{In terms of overhead, as shown in Figure~\ref{fig:gps}(c), both \texttt{Read} and \texttt{Write} primarily rely on two network paths: PCIe for local CPU--GPU communication and an RDMA fabric, typically InfiniBand, for remote CPU--GPU communication. Modern clusters provide sufficient bandwidth for parameter transfers over both paths. For instance, in a Hopper-architecture GPU cluster, PCIe and the RDMA fabric provide up to 63~GB/s and 50~GB/s, respectively, allowing GPS to complete parameter shifts within seconds, as evaluated in \S\ref{subsec:overheads}. Moreover, since \texttt{Write} is only required during training, we optimize the training backend to overlap \texttt{Write} with training in the subsequent RL step, further hiding its overhead from the critical path.}

\lhy{In terms of scalability, as the cluster grows, the numbers of GPUs and CPUs increase proportionally. Consequently, the number of role replicas, which consume GPUs, and the number of replicas managed by GPS, which consume CPUs, also increase proportionally. The deployment and communication patterns at a larger scale are therefore replicated across more replicas, rather than accumulating on existing replicas. Thus, the overall overhead does not increase with cluster size. We evaluate this scalability in \S\ref{subsec:overheads}.}

\subsection{Role-Shifting Mechanism}
\label{subsec:shift}

\lhy{Building on GPS, we can shift model parameters across roles through a unified interface. However, the shifting must also handle role-specific context beyond model parameters, such as optimizer states in training and CUDA Graphs in rollout.}

\lhy{Simply discarding such context can cause the shifted resource to suffer from cold starts, such as rebuilding CUDA Graphs, or even affect RL correctness, such as reinitializing optimizer states. At the other extreme, preserving the context of all possible roles avoids cold starts but substantially increases GPU memory consumption. For example, GPUs would need to retain both optimizer states and CUDA Graphs even when the corresponding roles are not active, reducing the resources available for RL execution.}

\lhy{To avoid the limitations of these two extremes, we design a role-shifting mechanism that manages role-specific context based on its data type. Specifically, we classify role context into different categories and apply the most appropriate action to each, such as discard, offload, or preserve. By tailoring these actions to each category, we minimize GPU memory consumption while reducing cold-start overhead.}

\begin{figure}
    \centering
    \includegraphics[width=\linewidth]{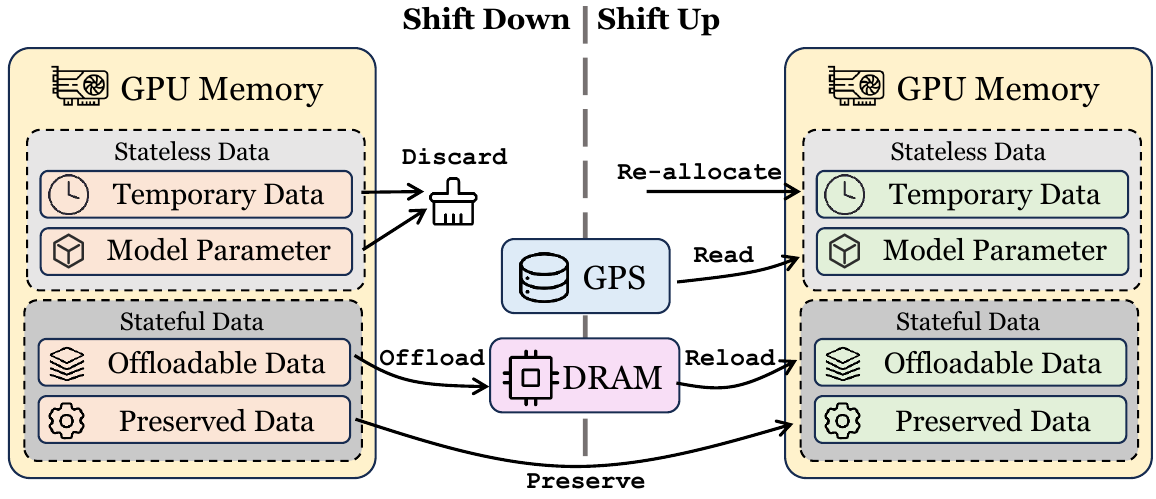}
    \myvspace{-20pt}
    \caption{\small{\lhy{The role-shifting mechanism. Role context is categorized into different types and handled accordingly. 
    }}}
    \label{fig:role-shifting mechanism}
    \myvspace{-10pt}
\end{figure}

\stitle{Categorizing role context.} 
\lhy{As shown in Figure~\ref{fig:role-shifting mechanism} (left), role context falls into two types: stateless and stateful.}

\lhy{Stateless data can be simply discarded when a role releases its GPUs. It consists of two types: (1) \textit{Temporary data}, such as KV Cache, activations, and temporary buffers. These data are only useful to the current role. When shifting roles, they can be directly discarded and rebuilt through recomputation. (2) \textit{Model parameters}, which can be treated as stateless data because all model weights are stored in GPS. They can therefore be discarded and \texttt{Read} from GPS at any time.}

\lhy{In contrast, stateful data must be retained rather than discarded to ensure correct RL execution and fast recovery after a resource shift. We further categorize stateful data into two types: (1) \textit{Offloadable data}, such as optimizer states, gradients, persistent training states, and CUDA Graphs. These data are required to preserve RL progress, but do not need to remain in GPU memory while another role is running. They can therefore be offloaded to DRAM (i.e., the local CPU memory of the GPU), and reloaded when needed. (2) \textit{Preserved data}, such as the CUDA context, NCCL communication groups, and several descriptions or definitions. 
These data are small enough (they remain less than 3.37 GB per GPU in all our experiments), so we preserve them on the GPU.}

\stitle{\textit{Shift down} and \textit{shift up}.} 
\lhy{Given this categorization, we split the shift into two standalone procedures: \textit{shift down} for the original role and \textit{shift up} for the new role.}

\lhy{As shown in Figure~\ref{fig:role-shifting mechanism} (left), during \textit{shift down}, stateless data, including temporary data and model parameters, are discarded. Stateful data are either offloaded to DRAM or preserved on the GPU. During \textit{shift up}, as shown in Figure~\ref{fig:role-shifting mechanism} (right), temporary data are lazily reconstructed at runtime, so the system only needs to reallocate the required GPU memory for them, such as the memory for KV Cache. Model parameters are obtained directly from GPS through the \texttt{Read} operation. For stateful data, data previously offloaded to DRAM are reloaded to the GPU, while data already preserved on the GPU require no further action.}

\stitle{Shifting granularity.}
\lhy{In principle, GPS supports resource shifting at arbitrary granularities, as \texttt{Read} can retrieve model parameters for any set of GPUs under any parallelism strategy. In practice, however, we impose a restriction: rollout and reward are shifted at the granularity of individual replicas, while training is shifted as a whole. This choice is motivated by two considerations: restricting the shifting granularity (1) reduces the search space during planning, and (2) avoids additional overhead that finer-grained shifting may introduce.}

\lhy{For rollout and reward, shifting at the replica granularity corresponds to scaling along the data parallelism dimension. Thus, shifting only requires rescheduling sequences without modifying model partitioning. For training, treating it as a monolithic unit avoids searching over training parallelism strategies and eliminates overheads such as re-partitioning optimizer states and establishing new communication groups when parallelism strategies change. Overall, this design strikes a practical balance between shifting flexibility and system overhead, while finer-grained shifting remains an interesting direction for future exploration.}

\myvspace{-10pt}
\section{\lhy{Shift-Aware Scheduler}}
\label{sec:rebalance}

\lhy{\pivotrl's runtime can shift resources across roles, but doing so raises an immediate question: once a role's resource allocation changes, how should work be rescheduled within the role? Rollout and reward shift at replica granularity, so a \textit{shift-up} replica may initially have no sequences, while a \textit{shift-down} replica may have remaining sequences that must be rescheduled to others. This requires migrating the sequences across replicas to maximize the role's aggregate throughput.}

\lhy{To support such rescheduling under resource shifting, we redesign the scheduler instead of treating each shift as an external, one-shot event. Our goal is to reduce rescheduling to the scheduler's native scheduling problem: the scheduler already has a basic dispatch capability, we can keep this capability as the core while adding shift-time handling.}

\lhy{We realize this redesign in two steps. We first develop a throughput-optimal scheduling mechanism, which consists of a shared waiting queue that continuously dispatches sequences across replicas and a model that estimates each replica's generation throughput, thereby maximizing aggregate throughput (\S\ref{subsec:model}). This mechanism naturally incorporates rescheduling: sequences that need to migrate due to a shift can simply be returned to the queue and dispatched through the same throughput-maximizing mechanism, allowing the workload to be seamlessly rescheduled (\S\ref{subsec:migration}).}

\subsection{Throughput-optimal Scheduling Mechanism}
\label{subsec:model}\label{subsec:queue}

\lhy{Within the scheduler, we maintain a shared waiting queue.\footnote{The queue can be implemented as a priority queue with customizable priorities. By default, sequences are prioritized in ascending order of sequence ID, consistent with veRL~\cite{verl}.} The scheduler continuously selects the first sequence in the queue and dispatches it to a replica.}

\lhy{To maximize the throughput, the scheduler estimates the throughput before and after dispatching the sequence.}
\lhy{Following prior work~\cite{distserve}, we build a throughput model to estimate the generation throughput $T_i$ of replica $i$ based on its number of tokens $N^{\text{Token}}_i$ and number of sequences $N^{\text{Seq}}_i$:}
\begin{equation*}
\small
    T_i = {N^{\text{Seq}}_i}/{(A \times N^{\text{Token}}_i + \max (B,\ C \times N^{\text{Seq}}_i) + D)}.
\end{equation*}
\lhy{Here, $A$, $B$, $C$, and $D$ are constants profiled and fitted offline in advance. The detailed derivation is provided in Appendix~\ref{app:throughput}.}

For a sequence of length $l$, the scheduler uses this throughput model to estimate the throughput gain of dispatching it to each replica $i$ with sufficient KV Cache capacity:
\begin{equation*}
\small
    \Delta T_i = \frac{N^{\text{Seq}}_i + 1}{A \times (N^{\text{Token}}_i +l) + \max (B,\ C \times (N^{\text{Seq}}_i+1)) + D}-T_i.
\end{equation*}
\lhy{The scheduler then dispatches the sequence to the eligible replica with the highest $\Delta T_i$. If no replica has sufficient KV Cache capacity, the sequence remains in the shared queue until capacity becomes available.}

\subsection{Shift-Time Sequence Migration}
\label{subsec:migration}

\begin{figure}
    \centering
    \includegraphics[width=\linewidth]{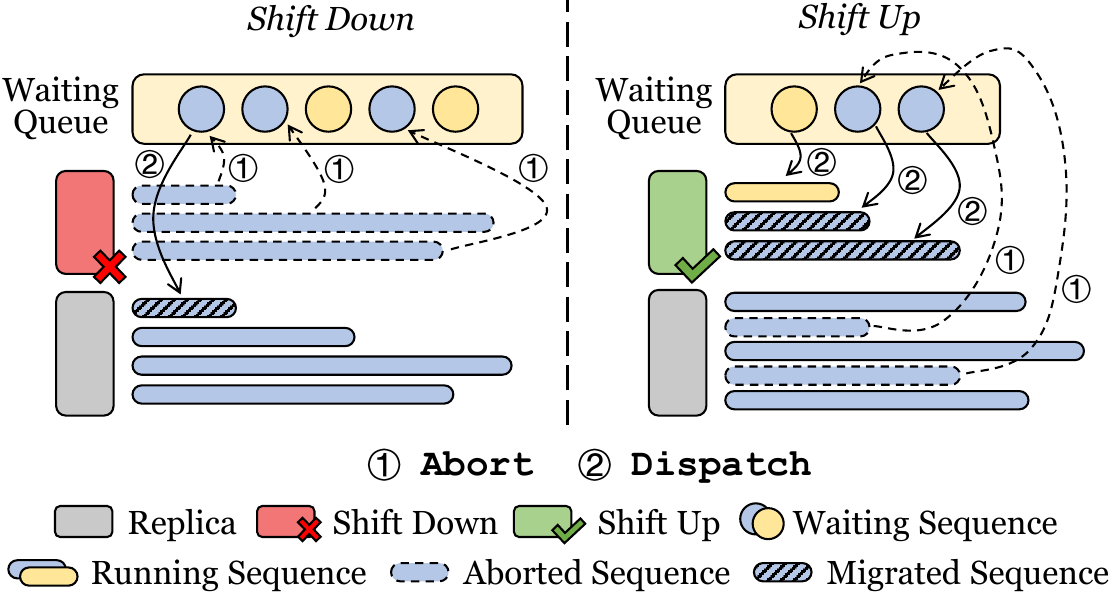}
    \myvspace{-20pt}
    \caption{\small{
    \lhy{Leveraging the shared waiting queue, sequence migration for both \textit{shift down} (left) and \textit{shift up} (right) requires only two operations: \texttt{Abort} and \texttt{Dispatch}.
    }}}
    \label{fig:throughput-optimal migration}
    \myvspace{-10pt}
\end{figure}

\lhy{As shown in Figure~\ref{fig:throughput-optimal migration}, by leveraging the shared waiting queue, sequence migration during the shift can be straightforwardly decomposed into two operations: (1) \texttt{Abort}, which selects sequences to abort from a replica and returns them to the shared waiting queue, and (2) \texttt{Dispatch}, which leverages the aforementioned throughput-maximizing dispatch mechanism to assign sequences to new replicas. We next detail how \textit{shift down} and \textit{shift up} perform sequence migration to reschedule the workload and improve throughput.}

\stitle{\textit{Shift down}.}
\lhy{As shown in Figure~\ref{fig:throughput-optimal migration} (left), the scheduler simply performs \texttt{Abort} on all sequences from the \textit{shift-down} replicas, returning them to the shared waiting queue. The scheduler then automatically performs \texttt{Dispatch} to assign these sequences to eligible replicas.}

\stitle{\textit{Shift up}.}
\lhy{As shown in Figure~\ref{fig:throughput-optimal migration} (right), a \textit{shift-up} replica starts with full capacity. We therefore first attempt to perform \texttt{Dispatch} for all sequences already waiting in the shared queue. If all waiting sequences can be dispatched, the remaining capacity indicates that the aggregate throughput may still benefit from migrating sequences from existing replicas. We then perform each migration as follows.}
\begin{itemize}[noitemsep, topsep=0pt, parsep=0pt, partopsep=0pt, leftmargin=*]
\item \lhy{We select the \textit{shortest sequence} from the replica with the \textit{highest estimated throughput} and perform \texttt{Abort} on it. This selection follows two considerations: (1) Throughput $T_i$ exhibits diminishing marginal gains as the number of sequences $N^{\text{Seq}}_i$ increases. Therefore, migrating a sequence from a high-throughput replica to a low-throughput replica can better improve aggregate throughput. (2) Shorter sequences incur lower migration costs because they require less KV Cache to be recomputed.}
\item \lhy{After each \texttt{Abort}, the sequence is immediately returned to the shared waiting queue, which then performs \texttt{Dispatch} to assign it to the most suitable replica.}
\end{itemize}
\lhy{We repeat this process until no further migration can improve the aggregate throughput across all replicas. A detailed pseudocode can be found in Appendix~\ref{subapp:mig_code}.}

\stitle{Complexity.}
\lhy{The upper-bound complexity of the migration procedure is $O\left(I\cdot N^{\text{Seq}} + I\log I\right)$, where $N^{\text{Seq}}$ stands for the total number of sequences among all replicas and $I$ stands for the number of replicas. Detailed analysis can be found in Appendix \ref{subapp:mig_complexity}.
In practice, the migration does not need to traverse all sequences, and its overhead is negligible, as evaluated by our experiments in \S\ref{subsec:overheads}.}

\section{\lhy{Planner}}
\label{sec:planner}

\begin{figure*}
    \centering
    \includegraphics[width=\linewidth]{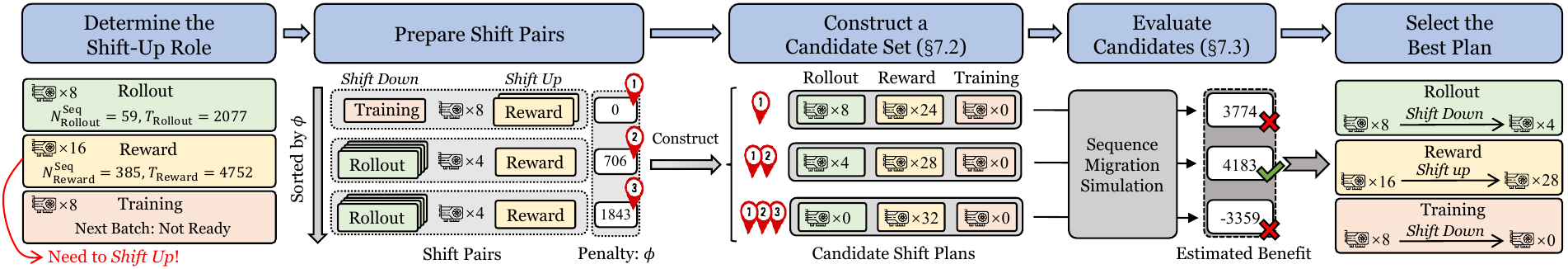}
    \myvspace{-22pt}
    \caption{\small{\lhy{The five-step procedure of the planner. In this example, each rollout replica holds 1 GPU, while each reward replica holds 4 GPUs. $N^\text{Seq}_\text{Role}$ denotes the total number of sequences processed by the role, $T_\text{Role}$ denotes its current throughput, and $\phi$ denotes the penalty.}}}
    \label{fig:decision procedure}
    \myvspace{-10pt}
\end{figure*}

\lhy{In this section, we introduce the planner that makes resource-shifting decisions. We first walk through our  overall procedure (\S\ref{subsec:planner overview}). Then, we detail its two main steps, candidate set construction (\S\ref{subsec:candidate condtruct}) and candidate evaluation (\S\ref{subsec:benefit evaluation}), to further illustrate how we make the best shift plan.}

\subsection{Five-step Procedure}
\label{subsec:planner overview}

\lhy{As illustrated in Figure~\ref{fig:decision procedure}, the planner follows five steps.}

\stitle{Determine the \textit{shift-up} role.}
\lhy{Given each role's workload state, including replica counts, running sequences $N^{\mathrm{Seq}}_{\text{Role}}$, throughput $T_{\text{Role}}$, training activity, and batch readiness, the planner first selects the role to \textit{shift up}. Specifically, if the next training batch is ready, training is selected. Otherwise, the planner selects between rollout and reward evaluation: the role with a larger generation latency $\tau_{\text{Role}}=N^{\mathrm{Seq}}_{\text{Role}}/T_{\text{Role}}$ is selected, as it has greater resource demand. In the example depicted in Figure~\ref{fig:decision procedure}, the next training batch is not ready, and $\tau_{\mathrm{Reward}}\approx0.08>\tau_{\mathrm{Rollout}}\approx0.03$, so reward is selected.}

\stitle{Prepare shift pairs.}
\lhy{Given the target role $r^{\uparrow}$ to \textit{shift up}, the planner identifies what to \textit{shift down} to release resources for $r^{\uparrow}$. This forms a set of shift pairs, each involving an exclusive set of GPUs. However, shifting is not free: once a replica starts to \textit{shift down}, the sequences running on it are stalled until the migration completes. We therefore assign each shift pair a penalty $\phi$, which captures the opportunity cost (i.e., the tokens that fail to be generated) during this shift period. Let $\tau_{\mathrm{shift}}$ denote the shift duration, which we treat as a constant, and let $m$ be the number of replicas that \textit{shift down} in the pair. Replica $i$ runs at a throughput of $T_{i}$, and thus forgoes $\tau_{\mathrm{shift}}\cdot T_{i}$ tokens. Summing over all gives $\phi = \sum_{i=1}^{m} \tau_{\mathrm{shift}} \cdot T_{i}$. The planner evaluates $\phi$ for every shift pair and sorts the pairs in ascending order, so that cheaper pairs are considered first. In the example, we have one 8-GPU training-to-reward shift pair and two 4-GPU rollout-to-reward shift pairs, listed in ascending order of penalty.}

\stitle{Construct a candidate set (\S\ref{subsec:candidate condtruct}).}
\lhy{Using the ordered shift pairs, the planner constructs a set $\mathcal{C}$ of feasible shift plans by combining one or more shift pairs with a dynamic programming algorithm. In the example, it produces three candidates.}

\stitle{Evaluate candidates (\S\ref{subsec:benefit evaluation}).}
\lhy{The planner estimates the benefit of each candidate through sequence migration simulation. 
In the example, the three candidates yield estimated benefits of $3774$, $4183$, and $-3359$.}

\stitle{Select the best plan.}
\lhy{Finally, the planner selects the candidate with the largest positive estimated benefit. If no candidate provides a positive benefit, the planner leaves the current resource allocation unchanged. In the example, the second candidate is selected, which will trigger a shift.}

\subsection{Candidate Set Construction}
\label{subsec:candidate condtruct}

\begin{algorithm}[t]
\small
\caption{\small{Candidate set construction}}
\label{alg:shift-plan-dp}
\begin{algorithmic}[1]
\Require Ordered shift pairs $\mathcal{P}$ and maximum
         \textit{shift-up} count $K$
\Ensure Shift plan candidate set $\mathcal{C}$

\State $D[0]\gets 0$; $D[1\ldots K]\gets+\infty$
\State $\mathcal{C}[0 \ldots K]\gets\varnothing$;

\ForAll{$P_i\in\mathcal{P}$}
    \State $\widehat{D}\gets D$;
           $\widehat{\mathcal{C}}\gets\mathcal{C}$
    \ForAll{$j$ such that $D[j]<+\infty$}
        \For{$e=1,\ldots,\min(r_i,K-j)$}
            \State $d\gets D[j]+\phi_i$;
                   $c\gets\mathcal{C}[j]\cup P_i^{(e)}$
            \State $\widehat{D}[j+e]\gets
                   \min\{\widehat{D}[j+e],d\}$
            \If{$d=\widehat{D}[j+e]$}
                \State Update shift plan
                       $\widehat{\mathcal{C}}[j+e]$ with $c$
            \EndIf
        \EndFor
    \EndFor
    \State $D\gets\widehat{D}$;
           $\mathcal{C}\gets\widehat{\mathcal{C}}$
\EndFor

\State \Return
$\{\mathcal{C}[k]\mid 1\leq k\leq K,\ D[k]<+\infty\}$
\end{algorithmic}
\end{algorithm}

\lhy{Since training performs \textit{shift up} as a whole, it does not introduce multiple candidates. We therefore focus on cases where rollout or reward need to \textit{shift up}.}

\stitle{Design rationale.}
\lhy{Suppose the \textit{shift-up} role $r^{\uparrow}$ owns $I$ replicas in total, of which $n$ are currently active. The planner must decide which of the remaining $K = I - n$ replicas to \textit{shift up}, so a naive enumeration over all subsets yields $2^{K}$ candidate plans, which is clearly impractical.}

\lhy{We prune this space using two intuitions. First, given a fixed workload, shift-time migration is deterministic (\S\ref{subsec:migration}), so $r^{\uparrow}$'s throughput after the shift depends only on how many replicas \textit{shift up} (denoted by $k$, $k \in [1, K]$), not on which ones. It therefore suffices to retain one candidate per $k$.}

\lhy{Second, given the \textit{shift-up} count $k$, selecting different replicas amounts to combining different shift pairs, and these pairs differ only in their penalty $\phi$. We therefore aim to select a set of shift pairs that provides exactly $k$ \textit{shift-up} replicas at the lowest total penalty (meaning achieving the same throughput at the lowest shift cost).}
\lhy{This can be formulated as a dynamic programming (DP) problem.}

\stitle{Dynamic programming.}
\lhy{As shown in Algorithm~\ref{alg:shift-plan-dp}, given the set of shift pairs $\mathcal{P}$, each pair $P_i \in \mathcal{P}$ contains $r_i$ ordered replicas of $r^{\uparrow}$. For each pair, we consider \textit{shifting up} the first $e$ replicas, where $e\in[0,r_i]$.}

\lhy{The DP state $D[j]$ stores the minimum penalty for \textit{shifting up} exactly $j$ replicas, while $\mathcal{C}[j]$ stores the corresponding shift plan. Initially, $D[0]=0$ and all other entries in $D$ are set to $\infty$, while all entries in $\mathcal{C}$ are set to $\varnothing$. For each shift pair $P_i$, the algorithm constructs a new DP table $\widehat{D}$ and $\widehat{\mathcal{C}}$. It then considers each reachable state $j$ and enumerates $e\leq\min(r_i,K-j)$ to extend $\mathcal{C}[j]$ by \textit{shifting up} the first $e$ replicas of $P_i$, denoted as $P_i^{(e)}$. The DP status only transfers when penalty $D[j]+\phi_i$ is lower than the current penalty $\widehat{D}[j+e]$, in which case $\widehat{\mathcal{C}}[j+e]$ is updated accordingly.}

The DP performs $O(KI)$ state transitions, yielding $O(I^2)$ time complexity since $K\leq I$. 
We evaluate the overhead of candidate set construction in \S\ref{subsec:overheads}, and refer interested readers to Appendix \ref{app:construction} for more implementation details.

\subsection{Candidate Benefit Evaluation}
\label{subsec:benefit evaluation}

\lhy{After the candidate set is constructed, we need to evaluate the benefit of each candidate to choose the best one.} 

\stitle{Overall throughput.} 
\lhy{Let $T_{\text{Rollout}}$ and $T_{\text{Reward}}$ denote the total throughput of all replicas in rollout and reward. We use their \textit{weighted harmonic mean} as the overall throughput:}
\begin{equation}
\small
\label{eq throughput}
\mathcal{T}
=
\left.
\left(
N_{\text{Rollout}}^{\text{Seq}} + N_{\text{Reward}}^{\text{Seq}}
\right)
\middle/
\left(
\frac{N_{\text{Rollout}}^{\text{Seq}}}{T_{\text{Rollout}}}
+
\frac{N_{\text{Reward}}^{\text{Seq}}}{T_{\text{Reward}}}
\right)
\right.,
\end{equation}
\lhy{where $N_{\text{Rollout}}^{\text{Seq}}$ and $N_{\text{Reward}}^{\text{Seq}}$ denote the total number of sequences for each role, including both sequences running on replicas and those waiting in the scheduler's queue (\S\ref{subsec:queue}).}

\stitle{Simulation for estimated throughput.} 
\lhy{To obtain $\mathcal{T}$ for a candidate plan, we need the precise workload state after shift-time sequence migration (\S\ref{subsec:migration}), so that we can use the throughput model in \S\ref{subsec:model} to estimate $T_{\text{Rollout}}$ and $T_{\text{Reward}}$. We therefore develop a simulator that replays the migration process using the algorithm in \S\ref{subsec:migration}. The simulator runs independently for each role of each candidate plan, allowing these simulations to execute in parallel with low overhead.}

\stitle{Final benefit.} 
\lhy{Let $\mathcal{H}$ be the time window over which the throughput improvement brought by a shift is expected to last (default as 60 seconds). A candidate that raises the system throughput from the current $\mathcal{T}_{\mathrm{current}}$ to $\mathcal{T}_{\mathrm{candidate}}$ therefore yields an expected token gain of $\mathcal{G} = \mathcal{H}\cdot(\mathcal{T}_{\mathrm{candidate}}-\mathcal{T}_{\mathrm{current}})$. Offsetting this gain by the penalty $\phi$ paid during the shift gives the final benefit of the candidate: $\mathcal{B} = \mathcal{G} - \phi$.}

\begin{figure*}[t]
    \centering
    \includegraphics[width=\linewidth]{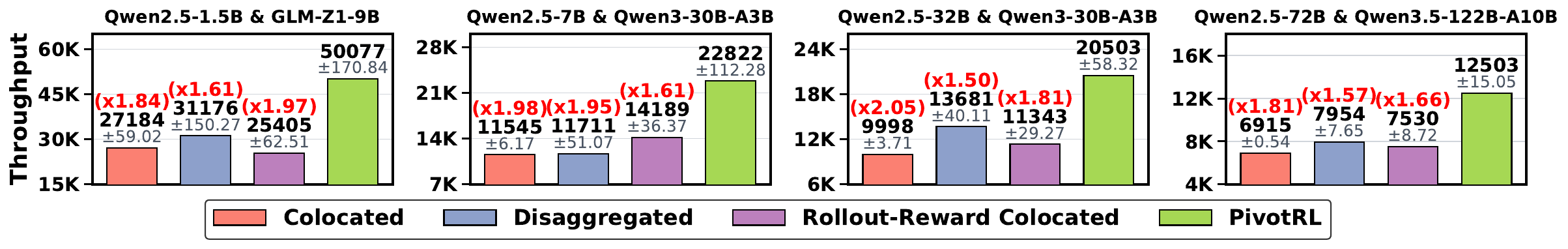}
    \myvspace{-25pt}
    \caption{\small{The end-to-end throughput of each deployment baseline and \pivotrl for different model groups. The values in parentheses indicate the speedup of \pivotrl over the corresponding baseline, while $\pm$ indicates the variation.}}
    \label{fig:throughput result}
    \myvspace{-10pt}
\end{figure*}
  
\section{Evaluation}
\lhy{
In this section, we evaluate \pivotrl with a focus on resource shifting. Our evaluation is guided by the following questions:
\begin{itemize}[noitemsep, topsep=0pt, parsep=0pt, partopsep=0pt, leftmargin=*]
\item Does resource-shiftable deployment in \pivotrl consistently provide throughput benefits over other deployment paradigms across different settings, and what is the magnitude of these benefits (\S\ref{subsec:e2e})?
\item How does \pivotrl shift resources among rollout, reward, and training to improve overall throughput (\S\ref{subsec:case})?
\item How do different shift strategies and system components affect the overall throughput of \pivotrl (\S\ref{subsec:ablation})?
\item How significant is the resource shifting overhead, and how does it scale with cluster size and model size (\S\ref{subsec:overheads})?
\item How do the hyperparameters introduced by resource shifting affect \pivotrl's performance (\S\ref{subsec:sens})?
\item Does resource shifting affect RL convergence (\S\ref{subsec:convergence})?
\end{itemize}
}

\subsection{Experiment Settings}
\stitle{Environment.} \lhy{We conduct all experiments on a cluster of 16 GPU nodes, each equipped with 8 NVIDIA H20 (96 GB) GPUs (128 GPUs in total). Intra-node GPUs are connected via NVLink with 900 GB/s bandwidth, and inter-node communication is over InfiniBand with 50 GB/s bandwidth.}

\stitle{Datasets and rewards.} 
\lhy{Following prior work~\cite{laminar, roll_flash, staleflow}, we use DAPO-Math-17k~\cite{dapo_dataset} as the training dataset and AIME-24~\cite{aime24} as the validation dataset, both of which are representative math reasoning datasets. During training, we use the sum of the verifiable reward and the GenRM reward as the reward for each trajectory, formulated as $R=R_{\text{DAPO}}+R_{\text{GenRM}}$. The prompt, policy response, and GenRM response are limited to 2K, 15K, and 20K tokens, respectively.}

\stitle{Models and metrics.} 
\lhy{To comprehensively evaluate \pivotrl across different policy models and GenRMs, we consider four groups of models, as shown in Table~\ref{tab:models}. These models cover different model families (Qwen~\cite{qwen_2, qwen_3} and GLM~\cite{glm5}), generations (Qwen2.5, Qwen3, and Qwen3.5), and architectures (dense and mixture-of-experts).}

\begin{table}[!t]
    \centering
    \small
    \caption{\small{The settings of each model group. TP: tensor parallelism, EP: expert parallelism.}}
    \myvspace{-10pt}
    \begin{tabular}{ccc}
    \toprule
        Policy Model & GenRM \\
    \midrule
        Qwen2.5-1.5B (TP=1) & GLM-Z1-9B (TP=1)\\
        Qwen2.5-7B (TP=1) & Qwen3-30B-A3B (TP, EP=4)\\
        Qwen2.5-32B (TP=4) & Qwen3-30B-A3B (TP, EP=4)\\
        Qwen2.5-72B (TP=8) & Qwen3.5-122B-A10B (TP, EP=8)\\
    \bottomrule
    \end{tabular}
    \myvspace{-10pt}
    \label{tab:models}
\end{table}

\stitle{Baselines.} 
\lhy{We select three baselines that represent the most common deployment paradigms in practice. (1) \textit{Colocated}, where rollout, reward, and training share the same GPUs across the entire cluster. 
(2) \textit{Disaggregated}, where rollout, reward, and training use separate GPUs. 
(3) \textit{Rollout-Reward colocated}, where rollout and reward share the same GPUs, while training uses isolated GPUs. 
}
\lhy{\pivotrl and all baselines are implemented on top of veRL~\cite{verl}, a state-of-the-art RL post-training system. We tune each baseline to its optimal configuration. Detailed GPU allocations for different roles in each baseline are provided in Appendix~\ref{app:deployment}.
}

\lhy{We evaluate system performance by end-to-end throughput, defined as the total tokens processed over several steps divided by the total time. Each experiment is repeated three times, and we report the mean and standard deviation.}

\begin{figure*}
    \centering
    \includegraphics[width=\linewidth]{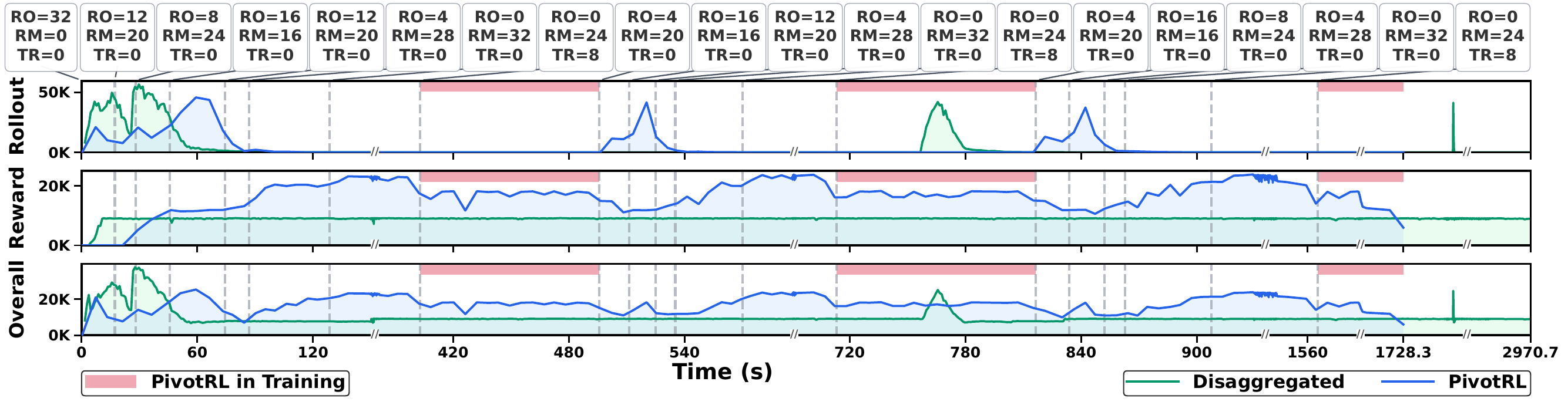}
    \myvspace{-23pt}
    \caption{\small{Throughput comparison between the disaggregated deployment and \pivotrl in Qwen2.5-7B \& Qwen3-30B-A3B. Top to bottom: rollout, reward, and overall throughput (Equation~\ref{eq throughput}). Resource shifts triggered by \pivotrl are marked, with the GPU allocations annotated for each role (RO: Rollout, RM: Reward, TR: Training). \lhy{The pink interval indicates that \pivotrl is in training (i.e., TR is not 0)}.}} 
    \label{fig:case-study-resource-shifts}
    \myvspace{-5pt}
\end{figure*}

\subsection{End-to-End Throughput}
\label{subsec:e2e}

\lhy{We first compare the end-to-end throughput with different model groups. As shown in Figure~\ref{fig:throughput result}, \pivotrl achieves the highest throughput among all deployment paradigms. (1) Compared to colocated deployment, \pivotrl pushes the throughput to 1.81--2.05$\times$. (2) Compared to disaggregated deployment, \pivotrl reaches a 1.5--1.95$\times$ throughput. (3) Compared to rollout-reward colocated deployment, \pivotrl also achieves 1.66--1.97$\times$ throughput.}

\lhy{The results can be explained as follows. (1) Colocated deployment suffers from severe throughput headroom at the end of each stage due to long-tail sequences, and since stages are not allowed to overlap, this headroom cannot be filled, making it the weakest among all settings.
(2) Disaggregated deployment improves performance by introducing asynchrony that allows different stages to overlap. However, the load of the three stages fluctuates, so resources are frequently mismatched with the actual demand and throughput headroom still appears. (3) Colocating rollout and reward partially inherits the advantages of the first two designs, and achieves higher throughput under the Qwen2.5-7B \& Qwen3-30B-A3B setting, but their drawbacks can also become more pronounced, as rollout-reward colocation amplifies the long-tail effect of rollout and reward. Compared with the baselines, \pivotrl performs the best by introducing resource shifting, which keeps the resource allocation aligned with the actual load and thus delivers the highest throughput.}

\subsection{Case Study}
\label{subsec:case}

\lhy{To gain insight into \pivotrl's behavior, we track the throughput and GPU allocation of both the disaggregated baseline and \pivotrl, as illustrated in Figure~\ref{fig:case-study-resource-shifts}.}

\lhy{For \pivotrl, we observe the following patterns: (1) At the beginning of the trace, \pivotrl assigns all resources to rollout, while reward has little work to process. As sequences accumulate, the bottleneck gradually moves downstream, prompting the planner to shift resources from rollout to reward. During this period, rollout throughput decreases as fewer replicas remain, while reward throughput increases. As a result, the overall throughput generally increases or remains high because the shifted GPUs provide greater marginal throughput when assigned to reward. (2) Once a training batch becomes ready, \pivotrl shifts GPUs to training. During this period, the overall throughput (Equation~\ref{eq throughput}) exhibits a drop because it accounts only for rollout and reward. Nevertheless, this shift is necessary to make progress on training and advance the RL step. (3) After training finishes, the training GPUs first shift back to rollout, followed by further gradual shifts from reward to rollout. This behavior occurs because a new step is unlocked, creating high resource demand at rollout. During this period, the overall throughput generally decreases because rollout has just started generating new sequences, while the downstream stages temporarily wait for these sequences to arrive.}

\lhy{In contrast, the disaggregated baseline maintains a static resource allocation, resulting in relatively stable throughput. However, it cannot adapt resources to the changing demands of different RL stages. Thus, although \pivotrl\ may experience temporary throughput fluctuations and can be slightly worse than the baseline in some periods, it achieves substantially higher overall throughput by dynamically adapting resources to evolving workload demands.}

\subsection{Ablation Study}
\label{subsec:ablation}

\lhy{To explain \pivotrl's throughput gains, we compare its planner's shift strategy against two heuristics and then analyze the contribution of each system component.}

\begin{figure}
    \centering
    \includegraphics[width=\linewidth]
    {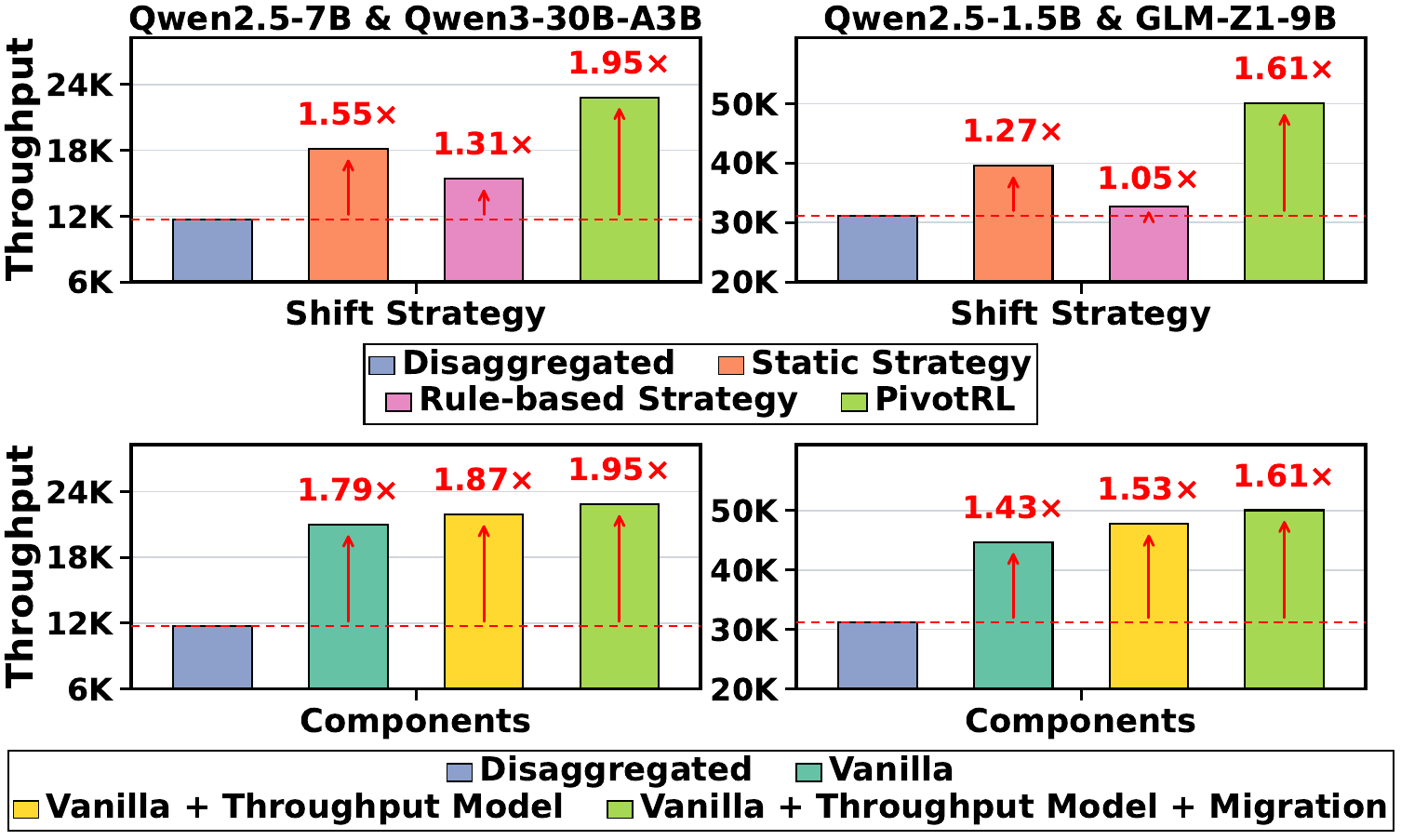}
    \myvspace{-22pt}
    \caption{\small{\lhy{Ablation study of shift strategies and system components across two model groups.}}} 
    \label{fig:ablation}
    \myvspace{-10pt}
\end{figure}

\stitle{Shift strategies.} 
\lhy{We compare \pivotrl with two simpler shift strategies. (1) \textit{Static strategy}, which only performs \textit{shift down} on training GPUs when the next batch is not ready and always splits them evenly between rollout and reward. 
(2) \textit{Rule-based strategy}, which reacts to local KV Cache utilization: a role performs \textit{shift up} when its utilization exceeds $0.8$ and \textit{shift down} when it falls below $0.1$.
As shown in Figure~\ref{fig:ablation} (top), under the two settings, the static strategy achieves $1.55\times$ and $1.27\times$ the throughput of disaggregated deployment, while the rule-based strategy achieves only $1.31\times$ and $1.05\times$. In contrast, \pivotrl achieves $1.95\times$ and $1.61\times$, respectively. This gap shows that neither a fixed split nor a local utilization signal accurately captures where resources are needed. Instead, comparing the aggregate throughput gain against the shift penalty, as \pivotrl does, can better reflect the value of a shift and guide more effective resource shifting.}

\stitle{System components.} \lhy{Figure~\ref{fig:ablation} (bottom) further examines how different components of \pivotrl contribute to its overall throughput. We first evaluate a vanilla variant that only enables the resource-shiftable runtime (\S\ref{sec:elasticity}) and planner (\S\ref{sec:planner}), without sequence migration or the throughput model for precise estimation (i.e., the throughput of a replica is treated as either 0 or 1, depending on whether it has a running sequence). It already improves throughput to $1.79\times$ and $1.43\times$. Incorporating the throughput model (\S\ref{subsec:model}) raises the gains to $1.87\times$ and $1.53\times$, while adding sequence migration (\S\ref{subsec:migration}) further reaches $1.95\times$ and $1.61\times$. Thus, resource shifting provides the largest gain, while the throughput model and sequence migration yield additional improvements by assigning workloads to resources more effectively.}

\subsection{Overhead and Scalability}
\label{subsec:overheads}

\lhy{We further analyze the overhead and scalability. Specifically, we measure the time spent on planning, performing \textit{shift up} and \textit{shift down}, and migrating sequences, while varying the GPU cluster size or model size to evaluate scalability.}

\stitle{Planner.}
\lhy{We divide the planning overhead into three parts. (1) \textit{Pre-processing}, where the planner analyzes the workload state to prepare shift pairs, (2) \textit{Candidate construction}, where the planner generates candidate shift plans (\S\ref{subsec:candidate condtruct}) (3) \textit{Simulation}, where the planner evaluates candidate benefits by simulating sequence migration (\S\ref{subsec:benefit evaluation}). Table~\ref{tab:planner-timing-homo} shows that the total planner overhead ranges from 0.31 to 2.95~s across different cluster sizes. Although the overhead of all three parts increases with the number of GPUs, it remains small. Moreover, the planner runs concurrently with RL execution, so its overhead does not affect system performance.}

\begin{table}[!t]
\footnotesize
  \centering
  \caption{\small{The time breakdown of the planner for Qwen2.5-1.5B \& GLM-Z1-9B and Qwen2.5-7B \& Qwen3-30B-A3B.}}
  \myvspace{-10pt}
  \label{tab:planner-timing-homo}
  \setlength{\tabcolsep}{5pt}
  \begin{tabular}{lcccc}
  \multicolumn{5}{c}{\textbf{Qwen2.5-1.5B \& GLM-Z1-9B}} \\
    \toprule
    GPU count & 16 & 32 & 64 & 128 \\
    \midrule
    Pre-processing (s) & \timing{0.06}{0.19} & \timing{0.18}{0.41} & \timing{0.47}{0.90} & \timing{0.39}{0.71} \\
    Construction (s)   & \timing{0.15}{0.14} & \timing{0.32}{0.28} & \timing{0.68}{0.53} & \timing{1.34}{1.21} \\
    Simulation (s)     & \timing{0.10}{0.13} & \timing{0.24}{0.17} & \timing{0.56}{0.52} & \timing{1.22}{0.91} \\
    \midrule
    Total (s)          & \timing{0.31}{0.27} & \timing{0.74}{0.52} & \timing{1.71}{1.17} & \timing{2.95}{1.67} \\
    \bottomrule
    \end{tabular}
    
    \begin{tabular}{lccc}
    \multicolumn{4}{c}{\rule{0pt}{8pt}\textbf{Qwen2.5-7B \& Qwen3-30B-A3B}} \\
    \toprule
    GPU count & 32 & 64 & 128 \\
    \midrule
    Pre-processing (s) & \timing{0.14}{0.38} & \timing{0.20}{0.55} & \timing{0.60}{0.78} \\
    Construction (s)   & \timing{0.14}{0.16} & \timing{0.31}{0.18} & \timing{0.84}{0.25} \\
    Simulation (s)     & \timing{0.27}{0.26} & \timing{0.65}{0.78} & \timing{1.37}{1.71} \\
    \midrule
    Total (s)          & \timing{0.55}{0.49} & \timing{1.16}{0.97} & \timing{2.81}{1.90} \\
    \bottomrule
    \end{tabular}
    \myvspace{-10pt}
\end{table}

\stitle{\textit{Shift down} and \textit{shift up}.} \lhy{We further measure the shift overhead across different model sizes and cluster sizes.}

\begin{figure}
    \centering
    \includegraphics[width=\linewidth]{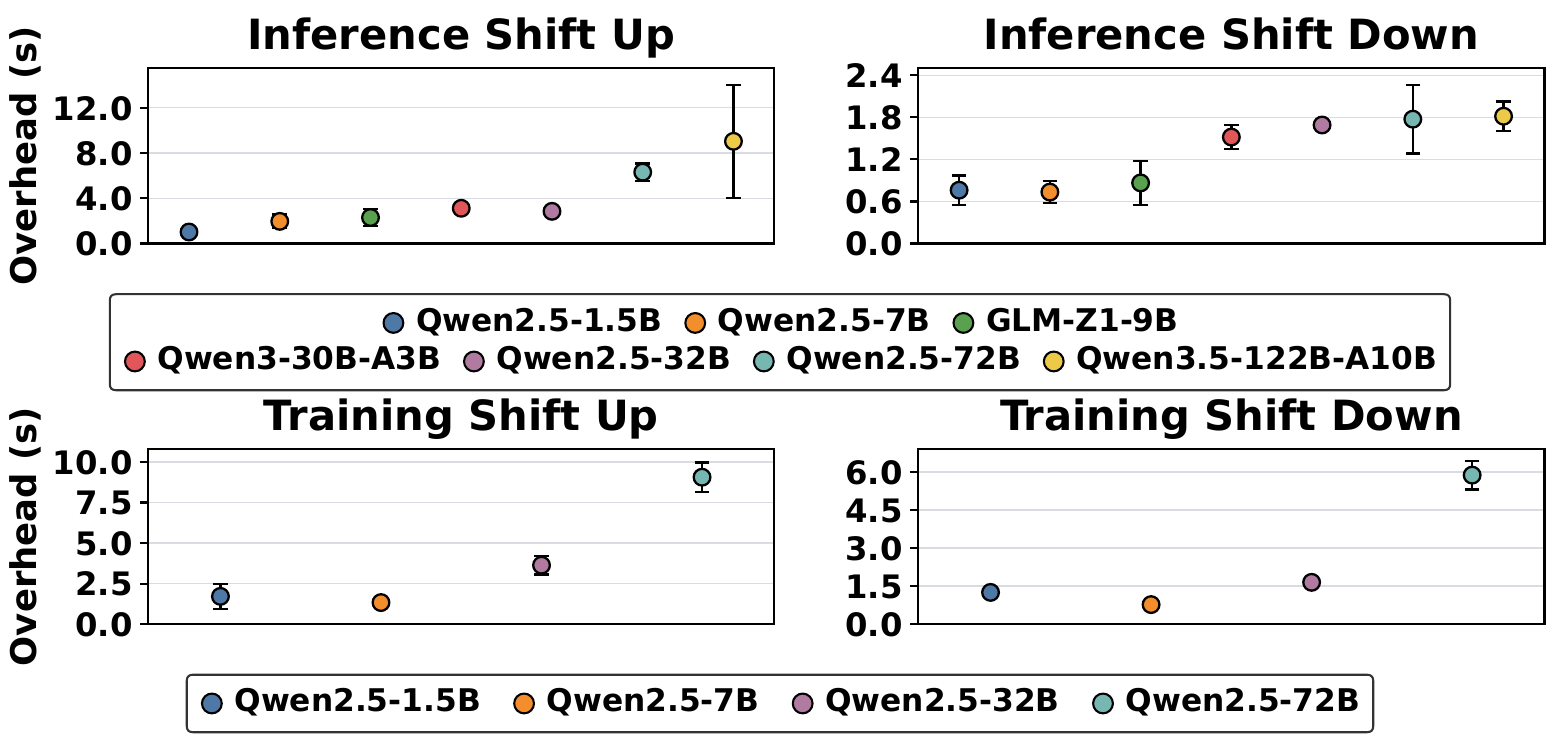}
        \myvspace{-26pt}
        \caption{\small{\lhy{Overhead of \textit{shift up} and \textit{shift down} for inference and training across different model sizes.}}}
        \label{fig:shift-latency-model-size}
        \myvspace{-10pt}
\end{figure}

\lhy{First, as shown in Figure~\ref{fig:shift-latency-model-size}, the overhead generally increases with model size. \textit{Shift up} is consistently more expensive because it needs to \texttt{Read} model parameters from GPS, whereas \textit{shift down} simply discards them. For inference (i.e., rollout and reward), the \textit{shift up} overhead increases from about 1~s for Qwen2.5-1.5B to about 9~s for Qwen3.5-122B-A10B, while \textit{shift down} remains below 2~s. For training, the \textit{shift up} overhead increases from about 2~s to 8.5~s, while \textit{shift down} remains below 6~s. The higher \textit{shift down} overhead for training is due to the additional context that must be offloaded (e.g., optimizer states).} 

\begin{figure}
    \centering
    \includegraphics[width=\linewidth]{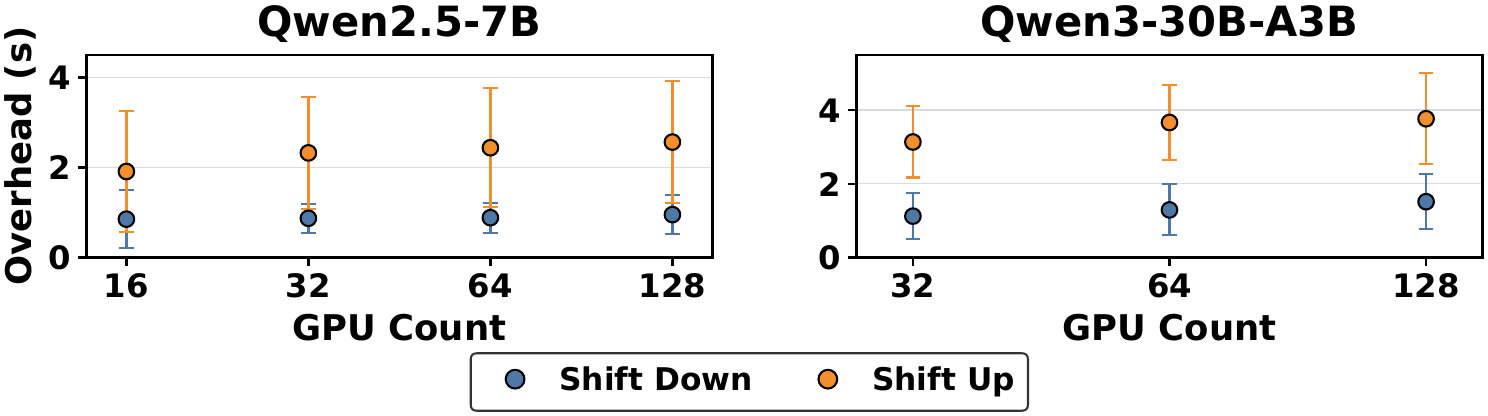}
        \myvspace{-22pt}
        \caption{\small{\lhy{Overhead of \textit{shift up} and \textit{shift down} for Qwen2.5-7B and Qwen3-30B-A3B under different GPU counts.}}}
        \label{fig:shift-latency-cluster-size}
        \myvspace{-10pt}
\end{figure}

\lhy{Second, as shown in Figure~\ref{fig:shift-latency-cluster-size}, increasing the cluster size causes only a modest increase in overhead. \textit{Shift up} remains within about 2--3~s for Qwen2.5-7B and 3--4~s for Qwen3-30B-A3B, while \textit{shift down} remains below 2~s.}

\lhy{Overall, both \textit{shift up} and \textit{shift down} complete within a few seconds, incurring minimal overhead.}

\stitle{Sequence Migration.} \lhy{We break down sequence migration overhead into three parts: (1) \texttt{Abort}, (2) \texttt{Dispatch}, including queue waiting time, and (3) re-prefill, where migrated sequences recompute their KV Cache before resuming generation. As shown in Table~\ref{tab:rebalance-timing-1p5b-glm9b}, \texttt{Dispatch} is the main source of overhead, taking 0.22--1.03~s, while \texttt{Abort} and re-prefill each take less than 0.19~s. The end-to-end overhead remains below 1.18~s across all cluster sizes. Moreover, migration only affects the execution of \textit{shift-up} replicas without pausing others, limiting its impact on overall performance.} 

\begin{table}[!t]
    \centering
    \caption{\small{\lhy{The per-sequence time breakdown during migration for Qwen2.5-1.5B \& GLM-Z1-9B and Qwen2.5-7B \& Qwen3-30B-A3B.}}}
    \myvspace{-10pt}
    \label{tab:rebalance-timing-1p5b-glm9b}
    \footnotesize
    \setlength{\tabcolsep}{5pt}
    \begin{tabular}{lcccc}
    \multicolumn{5}{c}{\textbf{Qwen2.5-1.5B \& GLM-Z1-9B}} \\
      \toprule
      GPU count & 16 & 32 & 64 & 128 \\
    \midrule
      \texttt{Abort} (s)  & \timing{0.07}{0.13} & \timing{0.07}{0.10} & \timing{0.10}{0.07} & \timing{0.12}{0.08} \\
      \texttt{Dispatch} (s) & \timing{0.72}{2.99} & \timing{1.03}{2.75} & \timing{0.53}{1.33} & \timing{0.55}{1.00} \\
      Re-prefill (s) & \timing{0.15}{0.24} & \timing{0.09}{0.08} & \timing{0.12}{0.10} & \timing{0.09}{0.05} \\
      \midrule
      Total (s)      & \timing{0.94}{3.06} & \timing{1.18}{2.79} & \timing{0.75}{1.36} & \timing{0.76}{1.01} \\
    \bottomrule
    \end{tabular}
    \begin{tabular}{lccc}
      \multicolumn{4}{c}{\rule{0pt}{8pt}\textbf{Qwen2.5-7B \& Qwen3-30B-A3B}} \\
      \toprule
      GPU count & 32 & 64 & 128 \\
      \midrule
      \texttt{Abort} (s)  & \timing{0.07}{0.08} & \timing{0.08}{0.05} & \timing{0.10}{0.07} \\
      \texttt{Dispatch} (s) & \timing{0.95}{2.23} & \timing{0.22}{0.62} & \timing{0.37}{0.66} \\
      Re-prefill (s) & \timing{0.09}{0.01} & \timing{0.13}{0.02} & \timing{0.19}{0.04} \\
      \midrule
      Total (s)      & \timing{1.12}{2.24} & \timing{0.43}{0.62} & \timing{0.67}{0.68} \\
      \bottomrule
    \end{tabular}
    \myvspace{-10pt}
  \end{table}

\subsection{Sensitivity Study}
\label{subsec:sens}

\lhy{Next, we investigate the sensitivity of end-to-end throughput to the only hyperparameter, the time window $\mathcal{H}$ (in seconds), which represents the expected duration over which a shift yields throughput benefits (see \S\ref{subsec:benefit evaluation}).}

\begin{figure}
\small
    \centering
    \includegraphics[width=\linewidth]{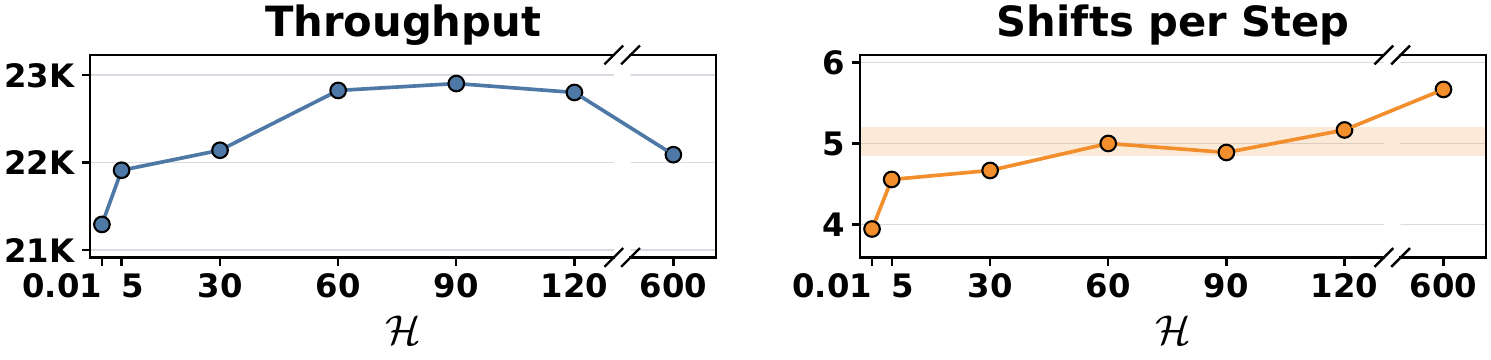}
    \myvspace{-25pt}
    \caption{\small{The throughput and average number of shifts taken per step with different $\mathcal{H}$.}}
    \label{fig:sensitivity}
    \myvspace{-5pt}
\end{figure}

\lhy{As shown in Figure \ref{fig:sensitivity}, we vary $\mathcal{H}$ from 0.01 to 600, with 60 as the default value used in the other experiments. For small $\mathcal{H}$, the expected throughput gains accumulated over this duration are often insufficient to offset the penalty $\phi$ associated with resource shifting. Consequently, the planner tends to be conservative in deciding when to shift, shifting resources only when the workload of a particular role is extremely light. This yields fewer shifts and lower throughput. In contrast, for excessively large $\mathcal{H}$, even marginal improvements in estimated throughput can accumulate enough benefit to outweigh $\phi$. This encourages overly aggressive resource shifting, leading to frequent shifts whose overhead ultimately reduces overall throughput. When $\mathcal{H}$ is set between 60 and 120, \pivotrl exhibits relatively stable performance in terms of throughput and number of shifts.}

\subsection{Convergence}
\label{subsec:convergence}

\begin{figure}
    \centering
    \includegraphics[width=\linewidth]{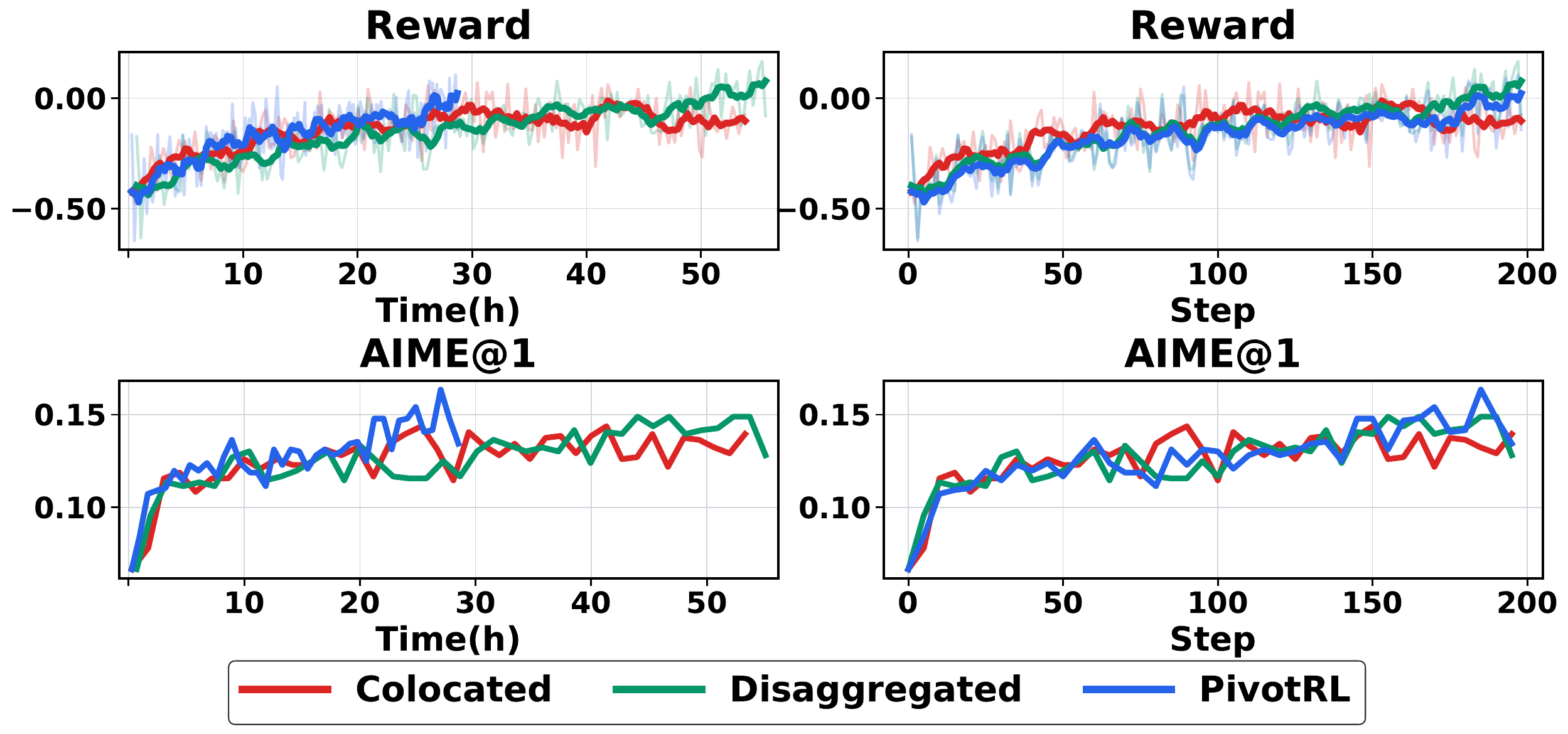}
    \myvspace{-20pt}
    \caption{\small{\lhy{RL post-training convergence of \pivotrl, colocated, and disaggregated deployments. Reward and AIME@1 are plotted against wall-clock time (left) and RL steps (right).}}}
    \label{fig:convergence}
    \myvspace{-10pt}
\end{figure}

\lhy{We compare the RL reward and AIME@1 of \pivotrl with the colocated and disaggregated deployments. As shown in Figure~\ref{fig:convergence}, the curves nearly overlap when plotted against RL steps, indicating comparable convergence. When plotted against wall-clock time, \pivotrl reaches the same reward and AIME@1 levels in about 30 hours, whereas the baselines require more than 50 hours. Thus, \pivotrl reaches comparable convergence in substantially less wall-clock time.}

\section{Related Works}
\stitle{RL post-training systems for LLMs.} Existing LLM RL post-training systems follow two deployment paradigms: colocated systems~\cite{verl,openrlhf,trl,slime} share GPUs across stages and execute them sequentially, while disaggregated systems~\cite{stream_rl, llama_rl, areal, roll_flash, rhy_rl, laminar, async_rlhf, asyncflow, staleflow} statically partition GPUs for each stage and overlap the stages. As discussed in \S\ref{sec:background}, both paradigms retain static resource allocations and cannot adapt to time-varying resource demands. 
We propose the resource-shiftable deployment paradigm to address this limitation.

\lhy{Recently, some concurrent work~\cite{dynarl, tiderl, bidirl} has also recognized the inefficiency of static resource allocation in RL with verifiable reward (RLVR), and dynamically adjusts the resources allocated to rollout and training. Unlike these works, our setting
further employs a GenRM, and therefore must shift resources across three roles spanning two models, which is considerably more challenging.}

Besides deployment, numerous works improve RL post-training efficiency in orthogonal ways: SortedRL, RollPacker, and Seer~\cite{sorted_rl,roll_packer,seer} mitigate long-tail generation through length-aware or fine-grained scheduling. Kimi k1.5, MiMo, AReaL, and AReaL-Hex~\cite{kimi_partial_rollout,mimo,areal,areal_hex} introduce partial rollout to break the synchronization barrier. Laminar and StaleFlow~\cite{laminar,staleflow} further exploit finer-grained per-replica asynchrony. 
These techniques are compatible with our work.

\stitle{Elasticity in LLM inference and training systems.} 
\lhy{A large body of work studies elasticity in LLM inference and training. On the inference side, Llumnix~\cite{llumnix} migrates requests across replicas, LoongServe~\cite{long_serve} varies the sequence-parallel degree for long-context requests, and SpotServe~\cite{spotserve} reconfigures parallelism and migrates state as GPU availability changes. On the training side, Pollux~\cite{pollux} jointly optimizes resource allocation and training efficiency, Singularity~\cite{singularity} transparently preempts, migrates, and resizes jobs, and Malleus~\cite{malleus} adapts hybrid parallelism to stragglers.
These systems, however, confine elasticity to a single, isolated inference or training stack, whereas RL couples rollout, reward, and training in one workload. Elasticity here entails shifting GPUs from one role to another, which is more complicated than scaling a single role up or down.}

\stitle{Reward-as-a-Service.} 
\lhy{Another line of work, including DistRS~\cite{distrs} and RollArt~\cite{rollart}, serves reward evaluation as a standalone service that supplies rewards to RL jobs in a multi-tenant manner. This approach, however, requires additional dedicated resources outside the cluster. More importantly, these systems target scenarios that are orthogonal to ours. Their rewards are often naturally amenable to cross-tenant sharing, such as rewards obtained through API calls or verification in sandboxes, and can therefore be provided as a shared microservice. In contrast, GenRMs are typically specific to individual RL jobs. This is particularly true for recent rubric-based GenRMs~\cite{rubrics, small_judge, alternate_rubric, train_rubric, rubricrm}, whose evaluation criteria are tightly coupled with the task and may need to be frequently updated or customized. Deploying GenRMs as a reward service is therefore not a good fit. Finally, a reward service does not change the fundamental problem: the mismatch between the static resource allocations of rollout
and training and their dynamic workload still remains, and resource shifting is what actually resolves this mismatch.}

\section{Conclusion}

\lhy{This paper studies RL post-training with generative reward models, where reward
evaluation itself becomes a full LLM inference workload and all three stages
contend for GPUs. We identified static resource allocation as a key bottleneck and proposed resource-shiftable deployment, realized in \pivotrl through a Global Parameter Storage with a unified role-shifting mechanism, a shift-aware scheduler, and a planner that decides when and how to shift. Our evaluation shows that \pivotrl improves end-to-end post-training throughput by up to $2.05\times$ across different models and cluster configurations, while preserving RL convergence.}

\bibliographystyle{ACM-Reference-Format}
\bibliography{reference}

\appendix
\clearpage
\section{Throughput Model Details}
\label{app:throughput}

\lhy{This appendix details the throughput model introduced in \S\ref{subsec:model}. This model is used by the scheduler to dispatch sequences (\S\ref{subsec:queue}), handle shift-time sequence migration (\S\ref{subsec:migration}), and used by the planner to estimate the benefit of a candidate shift plan (\S\ref{subsec:benefit evaluation}). Inspired by prior work~\cite{distserve}, we build a throughput model in which throughput is estimated from the number of running sequences and the total number of their tokens.}

\lhy{Most current inference systems batch active (running) sequences together and perform synchronous one-token-at-a-time generation (i.e., continuous batching~\cite{orca}). Therefore, we model the inter-token latency $L$ to compute generation throughput. For mainstream LLMs, the generation process can be approximated by considering only the attention and feed-forward network (FFN) layers.}

\stitle{Attention latency.}
\lhy{The latency of attention layers is dominated by memory accesses, specifically reads from and writes to the KV Cache. The rate at which data is loaded from the GPU's high-bandwidth memory (HBM) is generally determined by memory bandwidth. Consequently, the corresponding latency scales linearly with the KV Cache size.}
\lhy{Since the KV Cache size scales linearly with sequence length, attention latency can be modeled as a linear function of sequence length:}
\begin{equation*}
\small
    \tau_{\text{Attn}} = A \times N^{\text{Token}} + D_1,
\end{equation*}
\lhy{where $N^{\text{Token}}$ is the total length of all running sequences, and $A$ and $D_1$ are hardware-dependent constants.}

\stitle{FFN latency.} The latency of the FFN layers is determined by whichever is greater: the memory-access cost or the computation cost. FFN layers primarily perform matrix multiplications, so both matrix-loading time and matrix-multiplication time must be considered. The loading time can be modeled similarly to that of the attention layers. Because the data being loaded consists of model weights, this part of the latency is typically constant. Consider the multiplication of an $\mathbf{R}^{n\times h}$ matrix by an $\mathbf{R}^{h\times h'}$ matrix. The operation requires \(2nhh'\) floating-point operations, where \(n\) is the number of active sequences in the current decoding batch, and \(h\) and \(h'\) are the hidden dimensions. The computation time can therefore be modeled as a linear function of the number of active sequences. Together, the latency of FFN layers can be formulated as:
\begin{equation*}
\small
    \tau_{\text{FFN}} = \max (B,\ C \times N^\text{Seq}) +D_2,
\end{equation*}
where $N^{\text{Seq}}$ is the number of running sequences, and $B$, $C$, $D_2$ are constants determined by the hardware.

\stitle{Throughput.} As a conclusion, the inter-token latency of instance $i$ can be formulated as: 
\begin{equation*}
\small
    \tau_i=A \times N^\text{Token}_i + \max (B,\ C \times N^\text{Seq}_i) + D.
\end{equation*}
Further, the throughput of instance $i$ can be formulated as:
\begin{equation*}
\small
    T_i = \frac{N^\text{Seq}_i}{\tau_i}.
\end{equation*}
The constants $A,B,C,D$ can be profiled and fitted in advance, which is quite a fast procedure.

\begin{figure*}[!t]
\centering
\includegraphics[width=\linewidth]{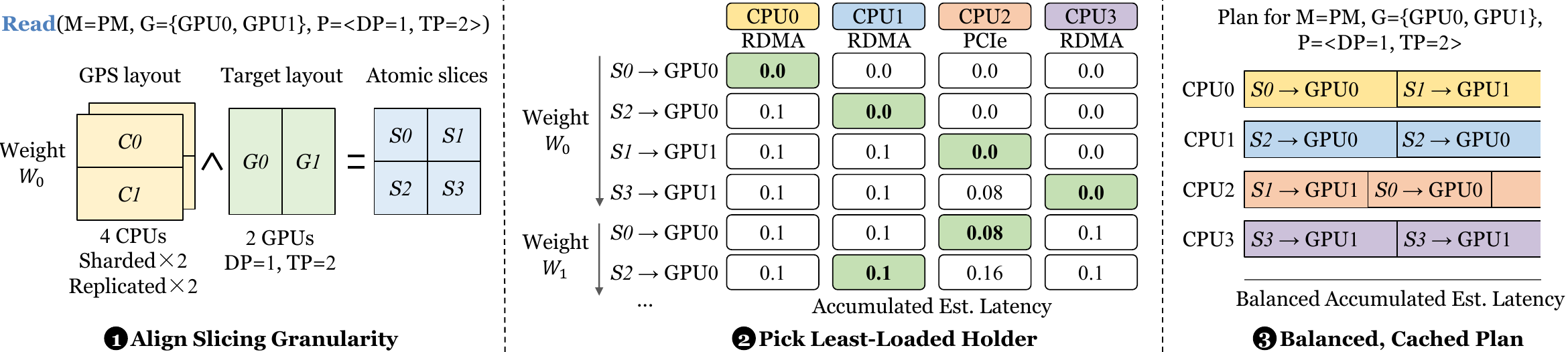}
\caption{\small{\lhy{The all-to-all collective underlying \texttt{Read} (\texttt{Write} is symmetric). In this example, GPU0 and GPU1 reside on CPU2, which can reach both GPUs over PCIe, while other CPUs must reach them via RDMA. \ding{182} The GPS layout and the caller's target layout are refined to their common granularity, yielding atomic slices $S_1$--$S_4$. \ding{183} We sweep the (slice, receiver) demands from top to bottom, assigning each to the candidate holder with the smallest accumulated estimated latency at that point (shaded). The cells show the accumulated latency at the time of each assignment. \ding{184} The resulting plan is balanced and cached for reuse.}}}
\label{fig:gps_a2a}
\end{figure*}

\section{GPS Details}
\label{app:gps}

\lhy{This appendix details the two engineering aspects of GPS deferred from \S\ref{subsec:ps}: how GPS partitions the parameters it holds in CPU memory (Appendix~\ref{app:gps_part}), and how the all-to-all collective underlying \texttt{Read} and \texttt{Write} is planned and executed (Appendix~\ref{app:gps_a2a}).}

\subsection{\lhy{Parameter Partitioning}}
\label{app:gps_part}

\stitle{\lhy{Read-Intensive workload.}}
\lhy{The partitioning of GPS determines which CPUs hold which parameters, and therefore how much traffic each \texttt{Read} and \texttt{Write} must move and over which links. Our choice is driven by the observation that resource shifting is a \emph{\texttt{Read}-intensive} workload. Within a single RL step, resources may be shifted several times, e.g., from reward back to rollout as the reward stage drains, and then from both to training once the batch is ready, and every such shift issues a \texttt{Read} on the newly assigned GPUs. In contrast, GPS is updated at most once per step: GenRM is frozen and thus never written, and the policy model is committed by exactly one \texttt{Write} after the training stage. \texttt{Read}s therefore outnumber \texttt{Write}s within a step.}

\lhy{The asymmetry is reinforced by where the two operations sit on the critical path. \texttt{Write} need only complete before the optimizer applies the next update, so it can be overlapped with the remainder of the training computation and is largely hidden. \texttt{Read}, on the other hand, gates the GPUs that have just performed \textit{shift up}: they cannot begin serving their new role until the parameters have landed. The non-overlapped cost of \texttt{Read} is thus the quantity that matters.}

\stitle{\lhy{Replication-First partitioning.}}
\lhy{Following this reasoning, GPS is \emph{fully replicated} whenever the memory budget permits: every node holds a complete copy of the parameters of all involved models in its CPU memory. Full replication makes every \texttt{Read} node-local. A \textit{shift-up} GPU can find all the slices it needs in the CPU memory of its own node, so the transfer is a CPU-to-GPU copy over PCIe and never touches the RDMA fabric. This is the fastest configuration available, and it also removes the fabric as a point of contention: concurrent \texttt{Read}s on different nodes proceed independently instead of competing for the same inter-node links. The price is paid by \texttt{Write}, which must now reach every replica and does traverse the fabric, but as argued above that cost is overlapped.}

\stitle{\lhy{Partitioning under memory pressure.}}
\lhy{Full replication requires the policy model and GenRM to fit together in the CPU memory of a single node. When they do not, we fall back to a hybrid layout analogous to HSDP~\cite{torchtitan}: the nodes of the cluster are divided into replication groups, the parameters are fully sharded \emph{within} each group so that every node holds only a $1/k$ fraction and the per-node footprint fits, and the groups are fully replicated \emph{across} the cluster. The group size $k$ is set to the smallest value that satisfies the memory constraint, which keeps the layout as close to full replication as the budget allows. Under this layout a \texttt{Read} still draws the bulk of its data from nearby nodes within one group, and only the $(k-1)/k$ remainder crosses the fabric. Besides, enlarging the cluster adds replication groups rather than widening the shard, so the \texttt{Read} cost does not grow with scale.}

\subsection{\lhy{All-to-All Collective}}
\label{app:gps_a2a}

\lhy{Because GPS and the caller partition each weight (i.e., parameter) differently, \texttt{Read} and \texttt{Write} entail re-partitioning data between a set of CPUs and a set of GPUs. We realize this re-partitioning as an all-to-all collective~\cite{hotspa, hydraulis, hetu}. The collective is executed through NIXL~\cite{nixl}. NIXL builds on UCX~\cite{ucx} and abstracts point-to-point transfers across CPU, GPU, and interconnect types behind a single interface. This abstraction allows GPS to be hosted on CPUs while communicating efficiently with GPUs in both directions. The remaining task is to decide which point-to-point transfers to issue. The \textit{communication plan} governs this decision. We describe the plan for \texttt{Read}, since \texttt{Write} is symmetric, with the roles of the CPU and GPU sides exchanged.}

\stitle{\lhy{Aligning slicing granularity.}}
\lhy{The two sides of the transfer do not agree on slice boundaries: a weight may be cut row-wise across GPS slices and column-wise by the target parallelism $P$, so a single GPS-side slice may be needed by several GPUs and a single GPU-side slice may have to be assembled from several GPS slices. As shown in step \ding{182} of Figure~\ref{fig:gps_a2a}, we first take the common refinement of the two layouts, which yields a set of \emph{atomic slices} that are indivisible on both sides. A converter performs this refinement, and in doing so also reconciles the differing parameter naming and layout conventions of the training backend, GPS, and the inference engine into one canonical representation. After refinement, every transfer moves a whole atomic slice, so no slice is ever split or concatenated in flight.}

\stitle{\lhy{Load-Balancing communication planning.}}
\lhy{Refinement turns the collective into a set of \emph{demands}, one per (atomic slice, receiving GPU) pair. Each demand admits multiple valid senders, since replication means an atomic slice generally resides on several CPUs, and any of them can serve it. Choosing arbitrarily concentrates traffic on a few senders and lets the slowest one dictate the completion time of the whole collective, so we choose deliberately. For every candidate sender we maintain an accumulated estimated latency, i.e., the total cost of the demands already assigned to it. The cost of one transfer is estimated as the size of the slice in bytes divided by the profiled bandwidth of the link between that sender and the receiver, plus a constant per-transfer overhead. Both terms are obtained by offline profiling, and the bandwidth term is what distinguishes a node-local PCIe path from a remote RDMA one. As illustrated in step \ding{183}, we walk the demands and greedily assign each to the candidate whose accumulated latency is currently smallest.}

\lhy{Step \ding{184} shows the effect. The greedy rule equalizes latency, so a sender on a fast link is given proportionally more data than one on a slow link, and the bottleneck sender is pushed down.}

\stitle{\lhy{Plan caching.}}
\lhy{The plan depends only on the signature $\langle M, G, P \rangle$ of the operation and on the profiled link bandwidths, none of which change between invocations with the same arguments. Since resource shifting may repeatedly return to the same configurations across RL steps, we compute each plan once and memorize it, so subsequent \texttt{Read}s and \texttt{Write}s with the same signature skip planning entirely and issue the cached set of point-to-point transfers directly.}

\section{Shift-Time Sequence Migration Details}
\label{app:migration}

\subsection{Pseudocode for Sequence Migration}
\label{subapp:mig_code}

We further describe the shift-time sequence migration procedure introduced in \S\ref{subsec:migration}. The procedure relies on two common operations, \texttt{Abort} and \texttt{Dispatch}, which are shared by both \textit{shift-down} and \textit{shift-up} migration.

\stitle{\texttt{Abort} and \texttt{Dispatch}.} 
\lhy{As shown in Algorithm~\ref{alg:migration-operations}, the \texttt{Abort} operation stops the selected sequences on a replica and inserts them into the shared waiting queue. The specification $d$ can select either all sequences on a replica or a particular sequence, such as the shortest one. The \texttt{Dispatch} operation repeatedly pops a sequence from the waiting queue, estimates the throughput gain $\Delta T_i$ of dispatching it to each replica (\S\ref{subsec:queue}), and assigns the sequence to the replica with the largest $\Delta T_i$.}

\begin{algorithm}[t]
\small
\caption{\small{Common operations for sequence migration}}
\label{alg:migration-operations}
\begin{algorithmic}[1]
\Function{Abort}{$r,d,Q$}
    \State \textcolor{commentblue}{\textit{// $r$: replica}}
    \State \textcolor{commentblue}{\textit{// $d$: selection specification, either \textsc{All} or a specific sequence}}
    \State \textcolor{commentblue}{\textit{// $Q$: shared waiting queue}}
    \State $S \gets r.\Call{GetSequences}{d}$
    \ForAll{$s \in S$}
        \State $r.\Call{Stop}{s}$
        \State $Q.\Call{Insert}{s}$
    \EndFor
\EndFunction
\Statex
\Function{Dispatch}{$Q,R$}
    \State \textcolor{commentblue}{\textit{// $Q$: shared waiting queue}}
    \State \textcolor{commentblue}{\textit{// $R$: set of candidate replicas}}
    \While{$Q \neq \emptyset$}
        \State $s \gets Q.\Call{Pop}{}$
        \ForAll{$r_i \in R$}
            \State Compute $\Delta T_i(s)$ as described in \S\ref{subsec:queue}
        \EndFor
        \State $r^\star \gets \arg\max_{r_i \in R}\Delta T_i(s)$
        \State $r^\star.\Call{Generate}{s}$
    \EndWhile
\EndFunction
\end{algorithmic}
\end{algorithm}

\stitle{\textit{Shift down}.} \lhy{When a set of replicas $R^{\downarrow}$ is scheduled to shift down, all sequences running on these replicas must be evacuated. As shown in Algorithm~\ref{alg:shift-down-migration}, \texttt{Abort} is called and all sequences are returned to the shared waiting queue $Q$. The scheduler then redispatches the aborted sequences among the remaining replicas $R'$. The sequence states are preserved during the abort process, allowing generation to resume from the previously produced tokens instead of restarting from scratch.}

\begin{algorithm}[t]
\small
\caption{\small{Sequence migration for \textit{shift down}}}
\label{alg:shift-down-migration}
\begin{algorithmic}[1]
\Function{ShiftDownMigration}{$R^{\downarrow},R',Q$}
    \State \textcolor{commentblue}{\textit{// $R^{\downarrow}$: replicas scheduled to shift down}}
    \State \textcolor{commentblue}{\textit{// $R'$: remaining replicas that can receive migrated sequences}}
    \State \textcolor{commentblue}{\textit{// $Q$: shared waiting queue}}
    \ForAll{$r \in R^{\downarrow}$}
        \State \Call{Abort}{$r,\textsc{All},Q$}
    \EndFor
    \State \Call{Dispatch}{$Q,R'$}
\EndFunction
\end{algorithmic}
\end{algorithm}

\stitle{\textit{Shift up}.} 
\lhy{As shown in Algorithm~\ref{alg:shift-up-migration}, for \textit{shift up}, the newly added replicas $R^{\uparrow}$ first consume the sequences already present in the waiting queue. However, this initial dispatch may still yield suboptimal throughput, since existing replicas can often host substantially more sequences. We therefore perform an iterative migration procedure over the entire replica set $R = R^{\uparrow} \cup R'$. In each iteration, the replica with the highest throughput is selected, and its shortest sequence is migrated. The procedure terminates when migration no longer improves the aggregate throughput.}

\begin{algorithm}[t]
\small
\caption{\small{Sequence migration for \textit{shift up}}}
\label{alg:shift-up-migration}
\begin{algorithmic}[1]
\Function{ShiftUpMigration}{$R^{\uparrow},R',Q$}
    \State \textcolor{commentblue}{\textit{// $R^{\uparrow}$: replicas scheduled to shift up}}
    \State \textcolor{commentblue}{\textit{// $R'$: existing replicas}}
    \State \textcolor{commentblue}{\textit{// $Q$: shared waiting queue}}
    \State \Call{Dispatch}{$Q,R^{\uparrow}$}
    \State $R \gets R^{\uparrow}\cup R'$
    \State $\mathcal{T} \gets \Call{EvaluateThroughput}{R}$
    \State Build a max-heap $h_M$ over $R$ keyed by replica throughput
    \While{\textbf{true}}
        \State $r_M \gets h_M.\Call{Pop}{}$
        \State \Call{Abort}{$r_M,\textsc{Shortest},Q$}
        \State \Call{Dispatch}{$Q,R$}
        \State $\mathcal{T}' \gets \Call{EvaluateThroughput}{R}$
        \If{$\mathcal{T}' \leq \mathcal{T}$}
            \State \textbf{break}
        \EndIf
        \State $\mathcal{T} \gets \mathcal{T}'$
        \State $\Call{UpdateThroughput}{r_M}$
        \State $h_M.\Call{Insert}{r_M}$
    \EndWhile
\EndFunction
\end{algorithmic}
\end{algorithm}

\subsection{Analysis of Complexity}
\label{subapp:mig_complexity}

The upper-bound complexity is $O\left(I\cdot N^{\text{Seq}} + I\log I\right)$, where $N^{\text{Seq}}$ denotes the total number of sequences across all replicas and $I$ denotes the number of replicas. 

\lhy{During \textit{shift down}, aborting sequences from the \textit{shift-down} replicas takes at most $O(I)$ time. Dispatching $N^{\text{Seq}}$ sequences across $I$ replicas takes $O(I\cdot N^{\text{Seq}})$ time. Thus, the overall complexity of sequence migration during \textit{shift down} is $O(I\cdot N^{\text{Seq}})$.}

During \textit{shift up}, the queued sequences are first dispatched to the newly activated replicas. The max-throughput replica is then selected iteratively for further sequence migration. The initial dispatch takes at most $O(I\cdot N^{\text{Seq}})$ time. Constructing the heap $h_M$ takes $O(I\log I)$ time. During subsequent dispatching, at most $N^{\text{Seq}}$ sequences are migrated across all replicas, which serves as an upper bound on the number of migration operations. For each migrated sequence, selecting its destination replica takes $O(I)$ time, while updating the corresponding replica in the heap takes $O(\log I)$ time. Thus, sequence migration takes $O(I\cdot N^{\text{Seq}})$ time, resulting in an overall complexity of
$O\left(I\cdot N^{\text{Seq}} + I\log I\right)$.

\begin{table*}[!t]
\small
    \centering
    \caption{\small{\lhy{Resource configurations for different model groups under various deployment paradigms. In the colocated paradigm, all three roles (rollout, reward, and training) share the same GPUs, so the total number is shown once across the three role columns. In the disaggregated paradigm, the numbers in the Rollout, Reward, and Training columns indicate the GPUs allocated to each role. In the rollout-reward colocated mode, the number spanning the Rollout and Reward columns denotes the GPUs shared by these two roles. For \pivotrl, since resources can be dynamically shifted across roles, each column shows the range of GPUs that can be allocated to the corresponding role. Since training is shifted as a whole, its allocation can only be zero or the full training workload.}}}
    \begin{tabular}{|c|c|c|c|c|c|}
    \hline
        \multirow{2}{*}{Policy Model} & \multirow{2}{*}{GenRM} & \multirow{2}{*}{Deployment} & \multicolumn{3}{c|}{GPUs for Each Role} \\
        \cline{4-6}
        & & & Rollout & Reward & Training \\
        \hline
        \multirow{4}{*}{\makecell[c]{Qwen2.5-1.5B \\ (TP=1)}}
&
\multirow{4}{*}{\makecell[c]{GLM-Z1-9B \\ (TP=1)}} & Colocated & \multicolumn{3}{c|}{32}\\
        \cline{3-6}
         & & Disaggregated & 8 & 8 & 16 \\
        \cline{3-6}
         & & Rollout-Reward Colocated & \multicolumn{2}{c|}{16} & 16 \\
        \cline{3-6}
         & & \pivotrl & 0--32 & 0--32 & 0 or 16 \\
        \hline
        \multirow{4}{*}{\makecell[c]{Qwen2.5-7B \\ (TP=1)}}
&
\multirow{4}{*}{\makecell[c]{Qwen3-30B-A3B \\ (TP, EP=4)}}  & Colocated & \multicolumn{3}{c|}{32}\\
        \cline{3-6}
         & & Disaggregated & 12 & 12 & 8 \\
        \cline{3-6}
         & & Rollout-Reward Colocated & \multicolumn{2}{c|}{24} & 8 \\
        \cline{3-6}
         & & \pivotrl & 0--32 & 0--32 & 0 or 8 \\
        \hline
        \multirow{4}{*}{\makecell[c]{Qwen2.5-32B \\ (TP=4)}}
&
\multirow{4}{*}{\makecell[c]{Qwen3-30B-A3B \\ (TP, EP=4)}} & Colocated & \multicolumn{3}{c|}{64}\\
        \cline{3-6}
         & & Disaggregated & 16 & 16 & 32 \\
        \cline{3-6}
         & & Rollout-Reward Colocated & \multicolumn{2}{c|}{32} & 32 \\
        \cline{3-6}
         & & \pivotrl & 0--64 & 0--64 & 0 or 32 \\
        \hline
        \multirow{4}{*}{\makecell[c]{Qwen2.5-72B \\ (TP=8)}}
&
\multirow{4}{*}{\makecell[c]{Qwen3.5-122B-A10B \\ (TP, EP=8)}}  & Colocated & \multicolumn{3}{c|}{128}\\
        \cline{3-6}
         & & Disaggregated & 32 & 32 & 64 \\
        \cline{3-6}
         & & Rollout-Reward Colocated & \multicolumn{2}{c|}{64} & 64 \\
        \cline{3-6}
         & & \pivotrl & 0--128 & 0--128 & 0 or 64 \\
        \hline
    \end{tabular}
    \label{tab:deployment}
\end{table*}

\section{Candidate Set Construction Details}
\label{app:construction}
\subsection{Shift Pair Representation}

\lhy{We consider rollout and reward, each serving a single model. Once the target role $r^{\uparrow}$ is selected, the planner constructs candidates from the ordered shift pairs $\mathcal{P}=(P_1,\ldots,P_g)$. Each shift pair $P_i$ contains $r_i$ ordered eligible replicas to \textit{shift up}, a common set of opposite-role replicas to \textit{shift down}, and the corresponding penalty $\phi_i$.}

\lhy{For $1\leq e\leq r_i$, $P_i^{(e)}$ denotes the partial shift plan that includes the first $e$ \textit{shift-up} replicas in $P_i$ and its common \textit{shift-down} set. Thus, selecting any positive number of replicas from $P_i$ incurs $\phi_i$ only once. Since the preprocessed shift pairs are disjoint, their penalties can be accumulated directly.}

\subsection{DP Transition}

The penalty array $D[j]$ stores the minimum accumulated penalty for \textit{shifting up} exactly $j$ replicas using the processed shift pairs. The plan array $\mathcal{C}[j]$ stores the corresponding shift plan candidates. Initially, only the empty plan is reachable:
\begin{equation*}
\small
D[0]=0,\qquad
\mathcal{C}[0]=\varnothing,
\end{equation*}
while $D[j]=+\infty$ and $\mathcal{C}[j]=\varnothing$ for $1\leq j\leq K$.

Before processing $P_i$, the planner initializes
$\widehat{D}\gets D$ and
$\widehat{\mathcal{C}}\gets\mathcal{C}$.
These assignments preserve the states that skip $P_i$. For every reachable count $j$, the planner considers
\begin{equation*}
\small
1\leq e\leq\min(r_i,K-j).
\end{equation*}
The extended plan and its penalty are
\begin{equation*}
\small
c=\mathcal{C}[j]\cup P_i^{(e)},
\qquad
d=D[j]+\phi_i.
\end{equation*}
The penalty transition is
\begin{equation*}
\small
\widehat{D}[j+e]
=
\min\left\{
\widehat{D}[j+e],
D[j]+\phi_i
\right\}.
\end{equation*}
If $d$ attains the updated value of $\widehat{D}[j+e]$, the planner updates $\widehat{\mathcal{C}}[j+e]$ with $c$. When penalties are equal, the later candidate in the deterministic traversal order is retained. After processing $P_i$, the next tables replace the current tables.

After all shift pairs have been processed, every reachable positive count contributes one candidate:
\begin{equation*}
\small
\mathcal{C}_{\mathrm{out}}
=
\left\{
\mathcal{C}[k]
\mid
1\leq k\leq K,\ D[k]<+\infty
\right\}.
\end{equation*}
The candidate set therefore contains at most $K$ shift plans.

\subsection{Complexity Analysis}

Let $g$ be the number of non-empty shift pairs, $I$ be the total number of replicas in the \textit{shift-up} role, and 
$K=\sum_{i=1}^{g}r_i$
be the number of eligible shift-up replicas. The upper-bound of $g$ is $I$, when the numbers of GPUs per each replica in both roles are the same. Copying the two DP tables for each shift pair requires $O(Kg)$ entry assignments. The number of non-skip transitions is bounded by
\begin{equation*}
\small
T_{\mathrm{DP}}
\leq
\sum_{i=1}^{g}
\sum_{j=0}^{K}
\min(r_i,K-j)
\leq
K\sum_{i=1}^{g}r_i
=
O(KI).
\end{equation*}
Since every shift pair is non-empty, $g\leq I$, and the table initialization and transitions together require $O(KI)$ DP operations. When $K\leq I$, this becomes $O(I^2)$.

\section{Deployment Details}
\label{app:deployment}

\lhy{In this appendix, we provide detailed descriptions of the implementation of each method and the resource configurations used in our experiments.}

\stitle{Implementation.} \pivotrl and all baselines are built upon veRL~\cite{verl}.

\textit{Colocated} directly builds on the veRL framework, with only the necessary modifications to adapt the reward function and datasets. Upon completion of each rollout, reward evaluation, or training phase, all resources are released and reassigned to the subsequent phase.

\textit{Disaggregated} uses the same training and inference backends as veRL. Each role is assigned an independent resource pool that cannot be used by other roles. In addition, disaggregated introduces staleness\footnote{Staleness refers to the maximum model version lag between the policy used to generate rollout trajectories and the policy being trained on those trajectories.} to enable asynchronous RL post-training. In all our experiments, we set the data staleness to 2.

\textit{Rollout-Reward collocated} builds upon the disaggregated implementation. It allows the rollout and reward evaluation stages to share the same set of resources, while the training stage uses a separate resource pool. Once rollout completes sampling a batch, reward evaluation is launched on the same resources to evaluate the generated samples, and the cycle repeats. We set the data staleness to 2, allowing rollout and reward evaluation to continue processing subsequent batches while training proceeds concurrently.

\pivotrl is also built upon the disaggregated implementation. To enable resource shifting among roles, it implements all key components described in \S\ref{sec:overview}, including a resource-shiftable runtime (\S\ref{sec:elasticity}), a shift-aware scheduler for rollout and reward (\S\ref{sec:rebalance}) and a planner (\S\ref{sec:planner}). Besides we also implement a monitor to capture the workload state for the planner, and an executor to execute shift plans generated by the planner. Unlike the baselines above, \pivotrl allows all three stages to utilize resources from the entire resource pool. We also set the data staleness to 2.

\stitle{Resource allocation.}
The resource configurations for different baselines over different model groups are listed in Table~\ref{tab:deployment}. In all experiments, we keep the total number of GPUs for each model group the same across all baselines. For each baseline, we tune its resource configuration to achieve the best end-to-end throughput.

\end{document}